\documentclass[twoside]{article}
\usepackage{ecj,palatino,epsfig,latexsym,natbib}
 \usepackage{apalike}
\usepackage{amsmath,amsfonts,amssymb,subfigure}
\usepackage{algorithm}
\usepackage{array}
\usepackage[caption=false,font=normalsize,labelfont=sf,textfont=sf]{subfig}
\usepackage{textcomp}
\usepackage{stfloats}
\usepackage{url}
\usepackage{verbatim}
\usepackage{graphicx}
\usepackage{cite}
\usepackage{booktabs}
\usepackage{multirow}
\usepackage[table,xcdraw]{xcolor}
\usepackage{colortbl} 
\usepackage{algorithm}
\usepackage{algpseudocode}
\makeatletter
\algrenewcommand\alglinenumber[1]{\scriptsize #1}
\makeatother

\begin{document}

\ecjHeader{x}{x}{xxx-xxx}{201X}{Minimalist Genetic Programming}{L. Trujillo}
\title{\bf Minimalist Genetic Programming}  

\author{\name{\bf Leonardo Trujillo} \hfill \addr{leonardo.trujillo@tectijuana.edu.mx}\\ 
        \addr{Tecnol\'ogico Nacional de M\'exico/IT de Tijuana,Tijuana, BC, Me xico}\\ 
        \addr{LASIGE, Department of Informatics, Faculty of Sciences, University of Lisbon, Lisbon, Portugal}
}

\maketitle

\begin{abstract}

Genetic programming (GP) is based on two important insights.
First, that any learning task can fundamentally be posed as a program induction problem, where
the goal is to construct a symbolic hierarchical model that is expressed as a syntax tree.
Second, to pose this task as a search problem, and use evolution to locate the desired model.
Since it was proposed, GP has produced notable results in a wide range of tasks and problem domains.
This work presents an alternative view by modifying the second core insight of GP,
posing the problem as a syntactic derivation task instead.
In particular, this paper presents Minimalist Genetic Programming (MGP), an algorithm that like
GP is biologically inspired, but instead of evolution it takes inspiration from
the Minimalist Program to human language, in which syntax is understood as an optimal solution
to the problem of linking two other mental systems.
In minimalism, the core computational process is a binary set formation operator called $MERGE$,
than can be used to incrementally construct complex syntactic structures using a simple
Markovian process.
MGP is able to discover the core building blocks of the symbolic expressions, and to
incrementally combined them using $MERGE$.
The proposed system is benchmarked on symbolic regression tasks that are known to be difficult to
solve with standard GP systems because of the propensity for bloat.
Results show that when a proper lexicon of atomic syntactic objects are chosen, MGP is able
to consistently produce the exact ground truth model on a set of symbolic regression problems where standard GP struggles to do the same.
The insights provided by minimalism are shown to be relevant to the problem of program induction,
and should be explored further based on the potential exhibited by MGP in this work.

\end{abstract}


\section{Introduction}\label{sec:intro}

In 1992 John Koza proposed Genetic Programming (GP), one of the core paradigms of evolutionary computation,
in his seminal book by the same name \citep{koza}.
Contextualized within the broader artificial intelligence (AI) research program, Koza made two key insights in his proposal.
To understand the first one, it is important to point out that AI, as a field of science, had been working under the conjecture
that ``every aspect of learning or any other feature of intelligence can
in principle be so precisely described that a machine [program] can be made to simulate it" \citep{mccarthy1955proposal}.
By the time of Koza's book most researchers in AI had concluded that precise descriptions of most ``features of intelligence" were proving extremely difficult to derive in a principled manner, let alone implement such theories in a ``machine" \citep{mitchell_artificial_2019}.
For that reason many had begun turning toward machine learning (ML) approaches, where models of intelligent behavior, and their implementations as programs, are not designed directly.
In ML these models are instead generated by way of a data-driven learning/search/optimization process.
The first key insight by Koza was that while most methods relied on an indirect representation for such models,
it might be prudent to explore the space computer programs directly.
Pointing out that ``for many problems, the most natural representation for a solution is a hierarchical computer program" (p. 63) \citep{koza};
i.e., many learning tasks can be posed as a problem of \textit{program induction}.
Koza's original version of GP is also known as tree-based GP, given that it directly encodes programs as hierarchical syntax trees.

Koza's second insight concerns the way in which this space of ``hierarchical programs" is explored to locate a program
with the desired behavior or performance.
This was the second key insight by Koza, to use an evolutionary process to search within this space.
Using evolution can be justified in several ways.
At the time of the book, there was already clear evidence that evolutionary computation, and in particular genetic algorithms,
could be used to implement powerful population-based global search methods.
These techniques are particularly useful in domains where
gradient information is not available \citep{Stork2020,Srensen2018}.
Moreover, it is well understood that evolution is capable of producing complex hierarchical and modular structures in biology \citep{Mengistu2016},
properties that computer programs also require.
The synergy between both of Koza's insights led to an impressive set of search and learning methods, that have solved a wide variety of problems
from diverse domains \citep{koza10}.
This paper focuses on one of the main application areas of GP, symbolic regression (SR) \citep{Kronberger2024,srbench2},
also one of the most intriguing applications of GP that was originally studied by Koza,
where the desired \textit{program} is a symbolic mathematical model that best fits a training dataset.

When comparing GP to other automatic modeling techniques, the fact that programs and models are expressed symbolically is probably
its most unique feature.
Such models are, at least potentially, intrinsically interpretable \citep{Atzmueller2024},
while most powerful black-box ML models are not \citep{rudin1}.
Interpretability allows a domain expert to gain valuable insights regarding the problem,
allowing them to extend and improve upon the solution generated by a GP search \citep{romera2024mathematical}.
However, while this potential is an in-built feature of GP, most current approaches often struggle to consistently achieve it \citep{Castelli2023}.
Similarly, while it is well understood that evolution often leads to modular and hierarchical structures, such modularity is difficult to
generate with GP \citep{gleam}.

The present work hypothesizes that a plausible reason for this unfulfilled potential lies with the manner
in which the space of programs is explored by GP.
One of the main obstacles to interpretability is related to bloat, the unnecessary increase in model size that does not lead to improvements in performance \citep{Silva2011}.
However bloat appears to be an in-built property of an artificial evolutionary process guided by a fitness function \citep{Langdon1998}.
While several heuristics can help mitigate it's effects \citep{perla},
the search for improved fitness will tend towards larger solutions over time.

Therefore, this paper proposes a new SR method that exploits Koza's first insight, but changes the proposal he made in the second.
This is not new, several works have used alternatives to evolution before, including using regularized regression \citep{ffx} or using
methodologies from industrial engineering \citep{kaizen}.
However, this work, like Koza's proposal, draws inspiration from biology, albeit from a different branch.
Instead of evolution, the proposal is to take inspiration from cognitive science,
and  develop a SR system inspired on the Minimalist Program to
Generative Grammar \citep{Chomsky1995,chomsky2004beyond,berwick2016why,komachi2019generative,Pan2024}.
The approach is called Minimalist Genetic Programming (MGP) (for obvious reasons), and is presented as a reformulation of what Koza proposed
over 30 years ago.
Instead of posing the task of program induction as a search problem, it is posed as a task of syntactic derivation.
While previous works have integrated grammars into GP systems before, these are hybrid approaches that use an evolutionary search that is constrained
by a context-free grammar \citep{ONeill2003,McKay2010}.
MGP is unique, replacing the elements of evolutionary theory to guide the model building process, with a formulation
based on minimalist syntax, a computational description of the core process that underlies the human capacity for language \citep{Chomsky1995}.

Before presenting MGP, several comments are pertinent to frame this research work.
First, unlike evolutionary theory, the Minimalist Program is a less mature theory, providing a conceptual framework to understand the human capacity for language at the computational and algorithmic level, but one which is still far from explaining the implementation level \citep{berwick2016why}.
However, the Minimalist Program does aim to cover all three levels of explanatory power, namely observational adequacy,
descriptive adequacy, and explanatory adequacy, while also providing a basis for exploring a possible solution to Darwin's problem\footnote{The apparent lack of selective advantage for the capacity for language.} \citep{darwin1871descent,berwick2016why}.
Second, like Koza's inspiration in evolutionary theory, our inspiration in the Minimalist Program is also sufficiently nuanced.
MGP, like GP, is not meant as a simulation of the biological process on which it is inspired,
instead it aims to apply the most relevant principles and core elements of the underlying biological theory,
while introducing necessary simplifications and non-biological modifications to solve the problem at hand.
Finally, one might argue that another bio-inspired computational system may be the last thing this research community needs \citep{Aranha2021}.
Critiques to the ever expanding literature on bio-inspired computing are mostly warrented and necessary \citep{Srensen2013}.
However, we believe that such critiques do not apply to MGP, since it introduces a biologically grounded and an algorithmically unique approach towards
solving the program induction problem.
Indeed MGP is not a meta-heuristic, in the sense of a global search algorithm as described in \citep{Stork2020}, it is a syntactic derivation system.
MGP, however, does include elements that are comparable to some of the core ideas in meta-heuristic search, such as exploration based on randomness and
exploitation based on performance, while also introducing ideas from related techniques,
like novelty search \citep{stanley2015greatness} and incremental evolution \citep{stanley2002evolving}.
While this paper is primarily concerned with motivating and outlining the main elements of a novel program induction approach based on MP,
MGP is benchmarked on standard SR tasks, exhibiting the capacity to reconstruct the exact ground truth mathematical expressions. 
MGP provides an alternative view to the program induction problem, with the potential to address some of the current challenges of traditional GP.
In the spirit of the original AI research program,
MGP takes inspiration from a theory of the core cognitive capacity that distinguishes humans from the rest
of the biological world \citep{darwin1871descent,berwick2016why},
offering a unique re-imagination of what it means to automate the construction of symbolic models.

The remainder of this paper is organized as follows.
Section \ref{sec:gp} presents an overview of GP, discussing the relevance of tree-structures in GP, evolution and cognitive science.
Section \ref{sec:minimalism} is intended as a brief, and simplified, introduction to the Minimalist Program.
Section \ref{sec:SR} outlines our proposal to use minimalist syntax to derive
SR models, and Section \ref{sec:MPSR} presents the proposed derivation system, called MGP.
Experiments and results are detailed in Section \ref{sec:experiments}.
Finally, Section \ref{sec:conclusions} contains a closing discussion,
conclusions and an overview of future work.

\section{Tree-based Genetic Programming}\label{sec:gp}
GP shares the basic algorithm structure of all global search techniques \citep{Stork2020}.
An iterative process that starts with a randomly generated set of solution candidates (population)
that are evaluated based on an objective (fitness) function\footnote{While the concept of fitness is specific to GP and evolutionary computation, in this work we will use it interchangeably with performance based on an objective or cost function.}.
A subset of solution candidates are then selected (based on fitness) to generate a new set of candidates,
some of which are kept (again based on fitness) to repeat the process until a termination criterion is met.
To  generate new solutions, evolutionary algorithms mostly employ unary or binary search operators, which are inspired on biological mutation and crossover.
These operators try to capture the idea of heredity, such that useful features from the current set of solutions are passed down over successive iterations (generations),
while novelty is introduced by an inherent amount of randomness in these operators.
Moreover, there are two unique features of GP, compared to other evolutionary approaches.
First, and most importantly, solution candidates are discrete symbolic structures that implement a computable expression, a program or a model.
Second, solution candidates do not have a fixed size or architecture, these aspects of the solutions are evolved along with their behavior.

In tree-based GP, as proposed by Koza, solution candidates are encoded using tree structures that
are constructed from a finite set of primitive elements, which include terminals (input variables, constants or 0-arity functions) and functions (operators of different kinds, that are expressed as n-arity functions), such that terminals are leaf nodes and functions are internal nodes.
Mutation and crossover are structural operations on trees, that commonly rely on randomly modifying or swapping sub-trees or nodes.

Many elements of the basic GP-based approach have been improved, extended or enhanced in different ways, while
other elements have largely been forgotten.
For instance, in the original GP formulation Koza proposed a concept called Automatically Defined Function (ADFs), the goal of which was to equip GP with a mechanism to discover and reuse modular structures.
However, research since then suggests that modularity is difficult to consistently evolve using GP \citep{gleam}, and most state-of-the-art methods do not explicitly promote or control modularity. 
One exception are multi-tree or  multi-expression representations \citep{m5gp,feat,ffx,gsgp}, which evolve a set of expressions that are then
linearly combined to construct the final model.
These strategies are normally used as feature transformation, or feature engineering methods.

While all programs can be expressed as trees, some GP-based approaches have argued that explicitly evolving trees is not necessarily the best approach.
Indeed, researchers have proposed representations based on lists \citep{linearGP}, stacks \citep{push} and graphs \citep{Miller2000CartesianGP},
to mention some of the alternatives.
Another approach has been to decouple the encoded model from the representation used within the search process.
For instance, Grammatical Evolution uses a string-based representation (similar to the one used in genetic algorithms)
that is transformed into a syntactic model through processing by a context-free grammar
\citep{ONeill2003}. 
These approaches change program representation, along with the required search operators,
but still follow an evolutionary and population-based approach to solve a search problem.

\subsection{Trees in GP and Beyond}
The explicit use of trees in the original formulation of GP can be justified beyond the fact that fundamentally all programs can be expressed as trees.

Hierarchical structures are common in biological systems, and evolution has shown a propensity
to evolve them \citep{Mengistu2016}.
Human design, in engineering and art, is also characterized by the use of hierarchic and modular structures \citep{KUPPURAJU1985}.
However, unlike the top-down hierarchies used in formal engineering, biological systems are built from the button up in
an incremental manner, an aggregative process that goes from simple to complex.
Similarly, the human mind has a tendency towards the construction and use of mental hierarchical structures \citep{berwick2016why,Fitch2014,Fitch2017}. 
Fitch describes this human ability as \textit{dendrophilia}, ``the human proclivity to attribute tree-like structures to sensory patterns ... in short, our \textit{love of trees}" \citep{Fitch2017}, and
states George Miller’s \textit{Supra-Regular Hypothesis} for human cognition as: ``when presented with a set of strings [data] humans have both a capacity and proclivity to infer hierarchical structures wherever possible" \citep{Fitch2014}.
This proclivity is informally evident in our scientific and philosophical understanding of reality, going back to the
Epicurians \citep{lucretius2007nature}.
Fitch is referring to very specific cognitive abilities, namely those underlying human language, and more precisely syntax, as well as music. This human capacity for language has been most precisely characterized by the seminal works of Noam Chomsky \citep{chomsky1965aspects}.
Recently, Berwick and Chomsky described human language as involving the manipulation of hierarchical structures, informally likened to “triangles [subtrees] in the mind" \citep{berwick2016why}.
Moreover, dendrophilia seems to be unique to humans \citep{Fitch2014,Fitch2017}, just like language \citep{berwick2016why}.
If this is correct, then it is clear that research in AI and ML should focus more on methods that explicitly construct and
manipulate tree-structures, to simulate at least some of the ``features of intelligence".
Fitch argues that the relevance of tree structures goes beyond the computational modeling of human syntax,
it can also be found at the implementation level of real neurons (not simplified models of neurons), with evidence
suggesting that characterizing their complex tree structure might be essential to understand their
full computational capabilities \citep{Fitch2014}.

The relationship with GP is clear, especially the GP of Koza.
Fitch is precise, defining a hierarchical structure as a ``structure whose graph takes the form of a rooted tree",
where a rooted tree is ``an acyclic, fully-connected graph with a designated root node" \citep{Fitch2014}.
Moreover, from the perspective of SR and model interpretability, dendrophilia suggests that
the best way to promote the construction if interpretable models may be by exploiting this proclivity explicitly.
Indeed, it does not seem like a coincidence that the most interpretable ML models are decision trees,
or that one of the most successful classes of models are built as ensembles of trees \citep{ShwartzZiv2022}.
We agree with Koza's proposal that the explicit use of tree-structures is well justified.
However, unlike Koza, we posit that the manner in which such structures are constructed need not follow a global search approach.
Inline with the original goals of the AI research program, this paper turns to what currently represents the most plausible
computational description of the core cognitive ability in humans that allows us to construct hierarchical
syntactic structures, the class of models that seem to be unique to the human mind \citep{Chomsky1995,chomsky2004beyond,berwick2016why,komachi2019generative,chomsky2023merge,Pan2024}.

\section{Overview of the Minimalist Program}\label{sec:minimalism}

Generative Grammar is a central framework in formal and theoretical linguistics, particularly in the study of syntax \citep{chomsky1957syntactic,chomsky1965aspects}.
It treats language as a biological and human system, and aims to answer the following questions:
\textit{How can there be just one human language and multiple languages at the same time?} \citep{chomsky2023merge}.
To answer this question, language is described as a formal system that generates, and can only generate, grammatically correct sentences of a language, in the idealized sense of linguistic competence.
Generative Grammar provides a computational description of language, backed up by strong empirical evidence \citep{berwick2016why}.
It has sharpened the obvious observation that humans, and only humans, are biologically endowed with a computational
system that allows the human child to acquire any one of many external languages.

Generative Grammar also provides an alternative approach to understand and describe high-level cognitive faculties,
in contrast to approaches based on classical AI.
While the field has taken many forms, we are interested in its most recent incarnation, the Minimalist Program (MP) \citep{Chomsky1995,Trotzke2020,chomsky2023merge,Pan2024}.
The MP describes language as an optimal and efficient computational system, based on a minimal set of computational mechanisms
and assumptions.
It advances the \textit{Strong Minimalist Thesis} (SMT), sating that language is an optimal solution to the problem of
connecting meaning and sound.
In other words, language, and in particular syntax, can be understood as a minimal computational system
that is both economical and optimally adapted to the interface requirements between two other mental systems:
(1) the Conceptual-Intentional (CI) system, from which thought, reasoning and meaning emerge and in which they are interpreted;
and (2) the sensorimotor (SM) system that handles speech, hearing and other forms of externalization and internalization.

A particularly intriguing possibility that emerges if the SMT is correct, is that it provides a tractable characterization of the human language faculty that can be studied using comparative biology to confirm whether it is unique to humans or not \citep{Fitch2014}. Moreover, it enables the development of scientifically grounded theories aimed at explaining how, and potentially when, the human capacity for language evolved \citep{berwick2016why}. 

\subsection{Core elements of the Minimalist Program}
This section provides core details of the main computational process in MP syntax,
following \citep{komachi2019generative,chomsky2023merge,Matsumoto2023,Pan2024}.
A core feature of language is that it can generate an infinite number of sentences using finite resources.
Therefore, we start from a finite set of atomic and irreducible elements called lexical items, contained in a Lexicon ($LEX$).
This is the raw material from which more complex syntactic objects ($SO$) can be built, such as sentences,
which are referred to as atomic items or atomic objects.
Lexical items have certain semantic features, rules that specify how they are semantically interpreted.
Moreover, there can be several general types or categories of items, and items of the same type will share similar semantic features.
However, the syntactic derivation process does not operate on the Lexicon directly, this is done using a Workspace ($WS$).
The workspace is a set of syntactic objects, and while two syntactic objects $SO_u$ and $SO_v$ may correspond to
semantically equivalent syntactic objects, they are treated as distinct within a workspace.
One may assume that the CI system is responsible with selecting which, and how many, lexical items to include at the start
of a derivation, with the goal of generating a grammatically correct sentence that can be sent to to the interfaces (CI and SM).
Between those systems, the main computational process used to build new syntactic objects is called $MERGE$, a binary set formation operator that is applied recursively to the workspace.
$MERGE$ is the process by which those ``triangles in the mind" are built, it represents the computational tool that constructs
the hierarchical structures that are characteristic of human language.

\subsection{Syntactic MERGE}
A Workspace $WS_i$ defines the stage of a syntactic derivation at time (or step) $i$, expressed as a set of accessible syntactic objects
$WS_i = \{ SO_a, SO_b, ... SO_x, SO_y \}$.
$MERGE$ is defined as an operator that performs binary set formation on $WS$, given by
\begin{equation}
   MERGE (SO_u, SO_v, WS_{i}) = WS_{i+1} \ , 
\end{equation} 
\noindent where $SO_u$ and $SO_v$ are accessible syntactic objects in $WS_{i}$.
The qualifier \textit{accessible} is crucial, differentiating between syntactic objects that can be used by $MERGE$, and those that cannot.
For instance, $WS_{i=0}$ only contains atomic and accessible syntactic objects.

For $i>0$ the workspace is a set of sets or a collection of sets, due to $MERGE$
progressively building nested hierarchical structures.
When $SO_u$ and $SO_v$ represent distinct elements in the workspace (i.e., $WS_i = \{ SO_a, SO_b, ... SO_u, SO_v \}$) the operator is referred to as External $MERGE$ ($EM$), given by
\begin{equation}
   EM (SO_u, SO_v, WS_{i}) = WS_{i+1} =  \{ SO_a, SO_b,... SO_{new} \} \ ,
\end{equation} 
\noindent where  $SO_{new} = \{ SO_u, SO_v \}$ represents the newly generated syntactic object.
Notice that $MERGE$, and in this case $EM$, not only introduces a new element but it also removes the syntactic objects that took
part in the operation; i.e., $SO_u$ and $SO_v$ are no longer in $WS_{i+1}$, they are now nested objects of $SO_{new}$.

However, after a $MERGE$ operation either $SO_u$ or $SO_v$, or both, may, under certain conditions, continue to be accessible;
i.e., they may still be used in future $MERGE$ operations, but only when $SO_{new}$ is also included in the operation. 
For instance, lets assume that $SO_v$ is still accessible in $SO_{new}$ ,
then a valid $MERGE$ operation called Internal $MERGE$ ($IM$) can be performed as
\begin{equation}
   IM (SO_{new}, SO_v, WS_{i+1}) = WS_{i+2} =  \{ SO_a, SO_b,...  SO_{new2} \} ,
\end{equation} 
\noindent where $SO_{new2} = \{ SO_v, \{ SO_u, SO_v \}\}$. In this case the lower copy of $S_v$ is no longer accessible after
an $IM$, but the upper copy may still be.

Notice that this is a Markovian process, with each step in the derivation only having access to the workspace from the previous step.
Neither $EM$ or $IM$ modify the syntactic objects on which they operate, $MERGE$ only assembles them
into a higher-order structures.
Furthermore, $MERGE$ either reduces the size of the workspace ($EM$) or keeps it the same ($IM$), in terms of the total number of distinct syntactic objects in the workspace.

\subsection{Derivation Process: Heads, Labeling, Minimal Search, Symmetry and Phase Transitions}
$MERGE$ is applied to the workspace until a single syntactic object is left, which is the final sentence sent to the interfaces.
In current MP the $MERGE$ operation is \textit{free}; i.e., it can be applied to any pair of accessible objects in the workspace, without any additional constraints.
In other words, $MERGE$  is not explicitly directed to combine any two particular objects in the workspace.

Lets turn to how, and when, syntactic objects are sent to the interfaces for semantic interpretation,
referred to as Transfer or Phase Transition.
At the start of the derivation, all syntactic objects are atoms, with unique properties.
Atoms have prespecified semantic interpretations or labels, which define the type of objects they are.
Then at each derivation step a new syntactic object is created by $MERGE$.
However, composite syntactic objects have to be \textit{labeled} in order for their semantic
content to be potentially interpreted at the interfaces, this is done by a Labeling Algorithm, by first detecting which of the two merged objects
will define the \textit{head}. The head object will influence how the new syntactic object is labeled.
Heads are \textit{found} by Minimal Search, which imposes economical constraints on the process.

In order to assign a label it is necessary to consider several possibilities of an object produced by $MERGE$.
\begin{itemize}
    \item One of the objects is an atomic item, but the other is not (it was produced by $MERGE$); for example when merging
    $SO_a$ with $SO_{w} = \{ SO_u, SO_v \}$, which generates $SO_{new} =\{ SO_a, \{ SO_u, SO_v \}\}$. In this case, Minimal Search dictates that the top element in the hierarchy ($SO_a$) defines the \textit{head} of $SO_{new}$ and is used to assign a label to the new object, while the other object ($\{ SO_u, SO_v \}$) represents the \textit{complement} of $SO_{new}$.
    Minimal Search \textit{finds} ($SO_a$) because it is more efficient to do so, than it is to find the nested objects
    in $SO_{w}$.
    \item Both merged objects $SO_u$ with $SO_{v}$ are labeled and of the same depth (for instance, both are atoms),
    this is referred to as a \textit{symmetric} $MERGE$ which can be resolved in one of two ways.
    The Labeling Algorithm may break the symmetry if the objects are of distinct types and their features are complimentary, such that a semantically meaningful label can be assigned to $SO_{new}$.
    On the other hand, it may be that both objects share the same features because they are of the same type. In this case, a process called
    Feature Sharing may assign a more general label to the new syntactic object, which allows it to be labeled and potentially interpreted at the interfaces.
\end{itemize}

\begin{figure*}
    \centering
    \includegraphics[width=0.95\textwidth]{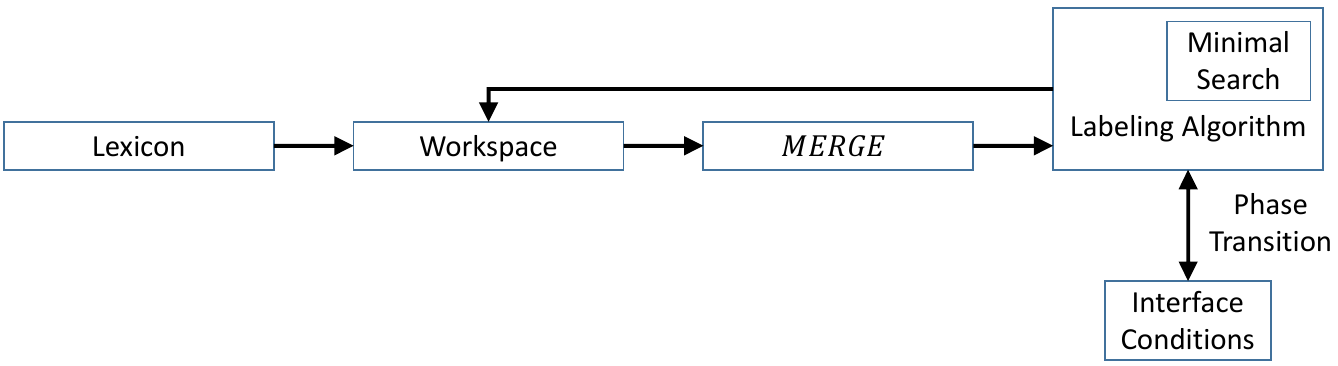}
    \caption{General overview of the main elements in MP syntax considered for this work.}
    \label{fig:mp}
\end{figure*}

To fully understand this process, some further details are necessary.
First, $MERGE$ does not trigger the Labeling Algorithm, it is only responsible for building the hierarchical structure (tree).
Labeling is triggered when Minimal Search can find the head of the new object, which means it can be interpreted in a meaningful way.
When the features of a head are fully specified by a particular $MERGE$ this constitutes a Phase Completion,
and now the syntactic object is referred to as a Phase, with a Phased Head and a Phase Complement.
In particular the complement of the Phase is \textit{sent} to the interfaces for externalization (SM) or semantic interpretation (CI).
However, the newly labeled Phase Head remains in the workspace, as if it is a new atomic lexical item with a unique interpretation,
and will be treated as such in subsequent $MERGE$ operations \citep{Trotzke2020}.
In this way higher-order semantic structures can be progressively constructed.
The above process continues until a single syntactic object remains in the workspace.
If the final object can be labeled and sent to the interfaces, this means that the derivation has converged;
if not, then the derivation is said to have crashed and the workspace is discarded.
This process is depicted clearly in Figure \ref{fig:mp}.

\begin{figure*}
    \centering
    \includegraphics[width=0.95\textwidth]{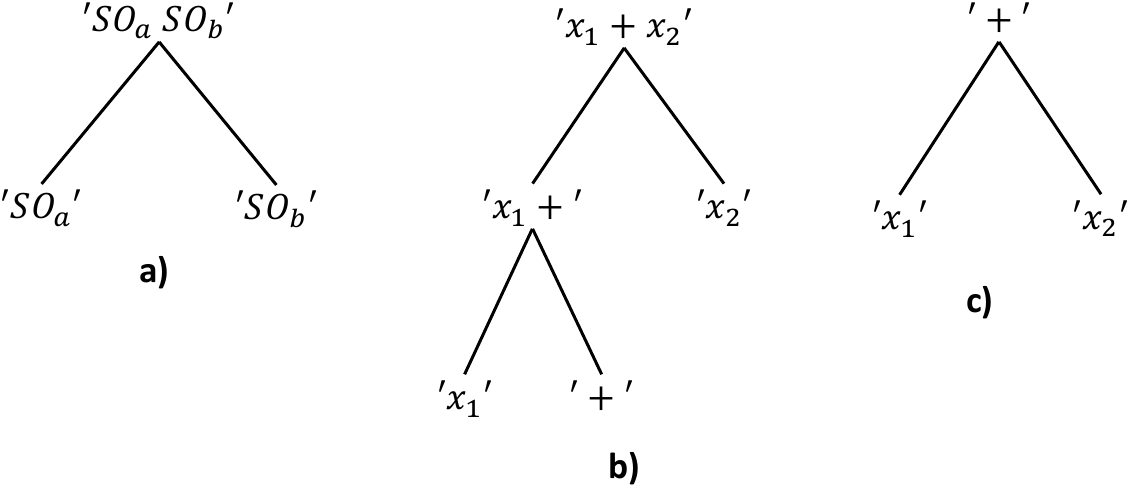}
    \caption{Trees generated by $MERGE$. (a) A generic tree generated by $MERGE$, $[_{LABEL}\ 'SO_a\ SO_b']$.
    (b) Example using a mathematical Lexicon $[_{semantic}\ 'x_1 + x_2']$.
    (C) The same expression using a typical GP tree.}
    \label{fig:trees}
\end{figure*}

One final note on notation and trees.
For example, if $MERGE$ is applied to $SO_a$ and $SO_{b}$, it produces $\{ SO_a, SO_b \}$. If this new object has a unique interpretation, the outcome of the Labeling Algorithm would be expressed as $[_{LABEL}\ 'SO_a\ SO_b']$.
Figure \ref{fig:trees}(a) illustrates this example in tree form.
Figure \ref{fig:trees}(b) shows a more realistic example for the SR domain, where $SO_a = [_{incomplete}\ 'x_1 +']$ and  $SO_b = [_{semantic}\ 'x_2']$, such that the newly merged object is $[_{semantic}\ 'x_1 + x_2']$ (the corresponding labels are explained in Section \ref{sec:MPSR}).
For reference, \ref{fig:trees}(c) shows the same example using a GP tree \citep{koza}.
Our notation for mathematical expression is a simplified one, but is sufficient for the purposes of this work \citep{Matsumoto2023}.

\subsection{Principles in the Minimalist Program}
The process described so far follows a set of important conceptual principles of MP syntax \citep{komachi2019generative,Pan2024},
some of which need to be emphasized.
First, there is a strong sense of Resource Restriction, starting from a finite workspace $MERGE$ never increases the number of syntactic objects (also called Minimal Yield), and it is strictly Markovian.
Second, the internal structure of syntactic objects cannot be modified by $MERGE$, their semantic interpretation is stable,
what is also referred to as the No-Tampering Condition or Phase Impenetrability Condition.
Third, economic constraints are also evident in Minimal Search for labeling, assuring that the most proximal element in a tree is used to label a structure, guaranteeing efficiency.
Fourth, $MERGE$ is \textit{free}, which means that it can be applied to any pair of accessible objects in the workspace.
This is, to say the least, somewhat counterintuitive since it leads to the labeling issues described above,
and can in principle generate syntactic gibberish.
It also seems suboptimal, with the derivation potentially \textit{wasting} time and effort.
However, it should not be interpreted as such. While $MERGE$ can be applied to any pair of objects, the process includes several
elements that allow the derivation process to handle problematic cases (symmetric $MERGE$), and interfaces to other systems (CI and SM) impose a final deterministic filter on the process that imposes constraints on the derivation.
Fifth, $MERGE$ is recursive and strictly binary, limiting the type of structures it can produce, but one that is consistent
with what is known of language \citep{Fitch2014,berwick2016why,Fitch2017}.

Finally, it can be reasonably speculated that after a syntactic object is sent to the CI system,
it may be stored as a single unit and \textit{repackaged} as an atomic element in future derivations.
This would allow for the Lexicon to be progressively expanded by reusing pre-constructed \textit{chunks} of syntax.
While this is speculative in MP, it has some theoretical plausibility \citep{marantz1997no,starke2010nanosyntax}.
In this work, we refer to this as a Multi-Derivational process, allowing us to account for the reuse of syntactic objects
that were constructed in earlier derivations within later ones, not just previous steps within the same derivation process.

\subsection{Contrasting MP with GP}
At this point, it is obvious that the starting point and the goal for both a MP syntactical derivation process
and a GP system are quite similar.
They both start from a set of primitive or atomic syntactic elements, called the Lexicon in MP and the Primitive set (terminals and functions) in GP.
The goal for both is to produce a semantically meaningful syntactic object, a sentence in MP and a program in GP,
both of which are hierarchical structures.

There are other interesting parallels despite their different internal mechanisms.
GP is a population-based global search method, which operates under the principles of exploration, exploitation and heredity, relying on random modifications and fitness based selection to guide a search.
On the other hand, MP derivations can be described as a rhythmic process \citep{Uriagereka2000},
that incrementally builds hierarchical structures.
This process also exploits randomness, or to be more consistent with MP, it uses a non-deterministic operator that
combines and recombines hierarchical structures.
Moreover, both approaches \textit{control} this randomness using a deterministic filtering mechanism,
selection and survival in GP, and interface conditions in MP.

Each method uses different operations to construct syntactic expressions. $MERGE$ in MP,
and the genetic operators (crossover and mutation) in GP.
$EM$ is akin to crossover, in the sense that it combines two
syntactic objects, while $IM$ is similar to mutation since it creates a new object from the syntactic material
contained in a single one.

GP progressively evolves a set of solution candidates, while MP incrementally builds a single structure.
In one sense, MP derivation is more akin to the tree generation methods used in GP,
like Grow or Full \citep{koza}.
However, these methods do not take into consideration any deterministic filtering of derived structures,
their goal is to introduce structural diversity
and hopefully discover useful building blocks for the subsequent evolutionary process.
Probably a more analogous conceptual process to GP would be the Multi-Derivation model we described before,
where multiple derivations are used to progressively construct increasingly complex syntactic objects.

Despite these parallels, there are also several key differences.
As was already mentioned, in terms of the tree building operators, $MERGE$ is exclusively constructive, while GP operators can be highly destructive.
While $MERGE$ can be applied freely, MP imposes Resource Restrictions and Minimal Yield constraints, which controls the complexity of the workspace and the size of derived structures.
On the other hand, GP is known to suffer from bloating, commonly evolving unnecessarily complex models \citep{Langdon1998}.
Another difference is that GP is intrinsically parallel, while MP describes a sequential and incremental process.
Finally, there is a strong sense of competition and selection pressure in GP.
\textit{Survival of the fittest} is one of the core ideas in the theory of evolution, with no parallel in MP.
From this perspective, it is reasonable to say that GP also follows the principle of Resource Restrictions, since there is only a limited amount of resources in the environment, this means that only some individuals can reproduce.


\section{Symbolic Regression using Minimalist Principles}
\label{sec:MPSR}
The goal of this section is to outline the first algorithmic proposal based on MP syntax to solve a program induction problem,
posing SR as a syntactic derivation task.
It is correct to say that MP syntax does not include a parallel for fitness or an explicit objective to guide
the derivation process, like fitness guides the search in GP, which is used for selecting good individuals and discarding poor ones.
Fitness in evolutionary algorithms, however, is implemented abstractly, since it is computed a priori, while in biological evolution
fitness is an a posteriori property of genes, not individuals \citep{dawkins1976selfish}.
Similarly, MP does offer a mechanism to discard useless derivations, imposed by third factor constraints at the interfaces
to other systems.
Like fitness in GP, these interface conditions can be implemented abstractly, where a problem-specific objective function may be used to select between
useful and not useful derivations.


\subsection{Implementation of $MERGE$ for Symbolic Regression}
Before presenting the proposed SR method based on MP syntax, which we call Minimalist GP or MGP,
it is necessary to discuss the solution representations and other technical details of how to implement the derivation process.
As discussed above, $MERGE$ is a tree-building operation, but it differs with those normally used in GP.
Figure \ref{fig:trees} illustrates the differences between trees in both paradigms.
In GP, functions or operators are encoded in internal nodes and terminal elements are leaves.
The design of initialization techniques, genetic operators, and protected function nodes in GP
guarantee that all generated trees represent valid and computable expressions.

The trees produced by $MERGE$ cannot guarantee this, since it is free to combine any accessible syntactic
object in the workspace.
However, MP derivations rely on other processes, those previously summarized in 
Figure \ref{fig:mp}.
It is important to mention that this depiction of MP derivations does not necessarily
follow all of the details of linguistic MP syntax.
Nonetheless, it provides the necessary tools to implement an MP-based derivation system
for SR.

Figures \ref{fig:trees2} - \ref{fig:trees4} present other instances of the types of trees that can constructed using $MERGE$.
For the explanation and examples that follow suppose that the initial workspace is given by $WS = \{ 'x_1', 'x_1', 'x_2', '+', '+', '\times', '-', '\div',sin()\}$, which was constructed from
Lexicon $LEX = \{ 'x_1','x_2', '+', '\times', '-', '\div','sin()'\}$; note that the workspace is constructed by sampling $LEX$ with replacement.
Like in GP, we have two types of atomic (primitive) elements in the workspace: (i) Functions or operators, namely $\{ '+', '\times', '-', '\div'\}$;  and (ii) Terminals or Variables, in this case only $\{ 'x_1'\}$.
For this work, we only consider unary and binary operators.

Terminal elements already represent potential solutions to a SR task, they are essentially naive predictors.
More generally in MP terms, we can say that a Phase Transition is triggered when a syntactic object has a meaningful interpretation,
particularly if it expresses a computable mathematical expression $K:\mathbb{R}^l \rightarrow \mathbb{R}$ that can be evaluated on a dataset
$D =  \{ (\vec{x_i},y_i\}$ where $\vec{x_i} \in \mathbb{R}^l$ such that $\vec{x_i} = (x_1, ..., x_l)_i$, $y_i \in \mathbb{R}$, $i=1,...,m$ and
$m$ is the number of samples in dataset and $l$ is the size of the feature space.
Each computable $K$ has a corresponding semantic vector defined as $S=(K(\vec{x_1}), ...,K(\vec{x_m}))$.
Such syntactic objects are labeled as \textit{semantic} objects, including terminal atoms like $[_{semantic}\ 'x_1\ ']$ or $ [_{semantic}\ 'x_2 \ ']$.

If an expression is not computable, the resulting object still has to be labeled for further derivation.
Atomic operator objects are labeled as such; i.e. $[_{operator}\ '+']$ or $ [_{operator}\ '\times']$.
Other objects created by $MERGE$ are labeled as follows.

\subsection{First order $MERGE$ between atomic syntactic objects}
\begin{figure*}
    \centering
    \includegraphics[width=0.95\textwidth]{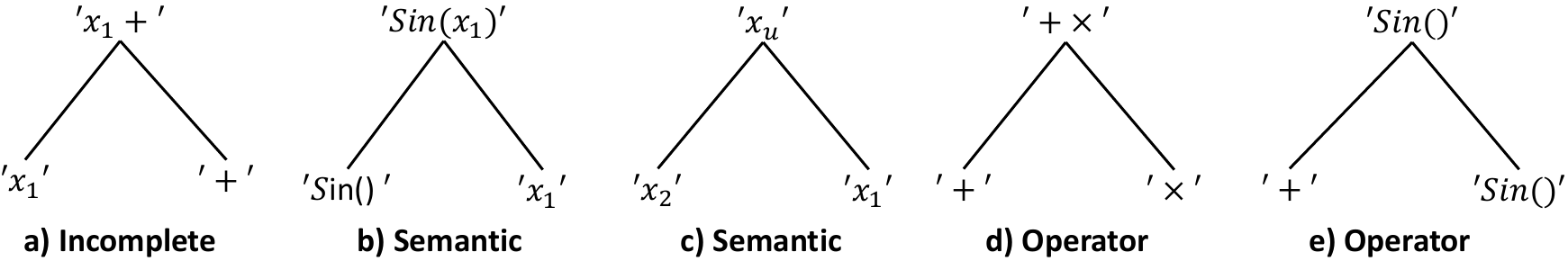}
    \caption{Examples of First-Order $MERGE$:
    (a) a binary operator and an atomic terminal $[_{incomplete}\ 'x_1 +']$;
    (b) a unary operator and an atomic terminal $[_{semantic}\ 'sin(x_1)']$;
    (c) two atomic terminals $x_1$ and $x_2$, labeled as $x_u\in{x_1,x_2}$ depending on the interface condition;
    (d) two operators with the same arity creates a list of operators $[_{operator}\ '+, \times']$; and
    (e) an unary operator $sin()$ and a binary operator $+$, labeled as $[_{operator}\ 'sin()']$.
    }
    \label{fig:trees2}
\end{figure*}

We begin with the first order $MERGE$ between two atomic objects, summarized in Figure \ref{fig:trees2}.
First, consider $MERGE$ of an \textit{operator} and a \textit{semantic} atomic object. For binary operators the new object is labeled as \textit{incomplete},
such as $[_{incomplete}\ 'x_1 +']$  or $[_{incomplete}\ 'x_1 \times']$, because they are not computable.
If the operator is unary (such as $sin()$ or $\sqrt{}$) $MERGE$ does trigger a Phase Transition and the object is evaluated,
computing its output semantics and labeling it as a \textit{semantic} object, such as $[_{semantic}\ 'sin(x_1)']$  or $[_{semantic}\ '\sqrt{x_2}']$.
 
Second, $MERGE$ of two \textit{semantic} objects, such as merging $[_{semantic}\ 'x_1\ ']$ with $[_{semantic}\ 'x_2\ ']$.
This is a symmetric $MERGE$, requiring a rule in the Labeling Algorithm to resolve the symmetry.
This is done by determining the head of the object based on a specified interface condition.
In this work, two possibilities are considered.
The first one is to use the objective function, such as an error measure $Error()$ to be minimized.
In this case, the new object will be labeled as $[_{semantic}\ 'x_1\ ']$ if $Error('x_1\ ',y) \leq Error('x_2\ ',y)$ with $y$ the target semantics, or vice versa. Another option is to determine the head of the object randomly.

Finally, we consider $MERGE$ of two \textit{operator} objects. This is another symmetric $MERGE$ that does not trigger a Phase Transition since it cannot specify a computable expression; these are also labeled as an \textit{operator} object.
For this case, we propose the following set of rules to determine the head of the object.
\begin{enumerate}
    \item $MERGE$ of operators with different arity. In this case, we resolve the symmetry by discarding the operator with the largest arity.
    For instance, if $'+'$ is merged with $'sin()'$ the resulting object would be $[_{operator}\ 'sin()']$.
    Since $IM$ is not used in this proposal, without this rule objects would tend to accumulate large portions of non-computable structures.
    \item $MERGE$ of operators with the same arity. This leads to the creation of a list of operators, such as $[_{operator}\ '+, \times']$  or $[_{operator}\ 'sin(), \sqrt{}']$. The way in which such objects are used in subsequent derivation steps is covered next.
\end{enumerate}

\subsection{Higher-order $MERGE$ using \textit{operator} objects}
\begin{figure*}
    \centering
    \includegraphics[width=0.95\textwidth]{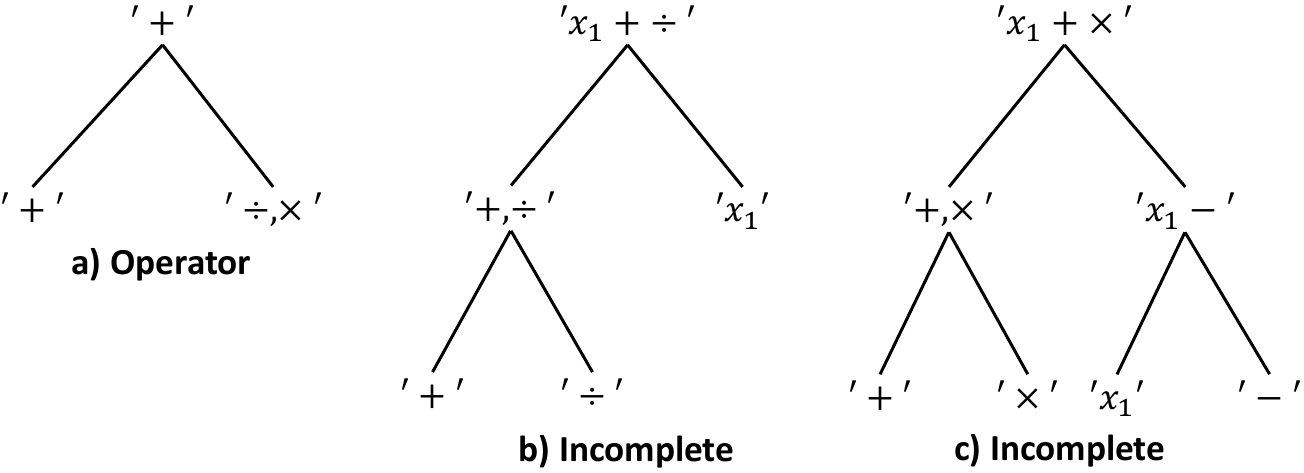}
    \caption{Examples of Higher-Order $MERGE$ with operator objects:
    (a)  non-symmetric merge between $[_{operator}\ '+']$ and $[_{operator}\ '\div','\times']$ to generate $[_{operator}\ '+']$;
    (b)  an operator object $[_{operator}\ '+,\div']$ and a semantic object $[_{semantic}\ 'x_1']$ to generate $[_{incomplete}\ 'x_1 \ +,\div']$; and
    (c)  an operator object $[_{operator}\ '+,\times']$ and an incomplete object $[_{incomplete}\ 'x_1 -']$ to generate $[_{incomplete}\ 'x_1 + \times']$.
    }
    \label{fig:trees3}
\end{figure*}

Consider $MERGE$ using arbitrary operator objects, with examples shown in Figure \ref{fig:trees3}.
First, consider when $MERGE$ is applied using two \textit{operator} objects, this can produce:
\begin{itemize}
    \item A symmetric $MERGE$, like $SO_{new} = \{SO_{O1}, SO_{O2} \}$ with
    $SO_{O1} = [_{operator}\ '+,\times']$ and $SO_{O2} = [_{operator}\ '\div,-']$, or other similar pairs of nested objects. 
    In this example, the resulting object would be $[_{operator}\ '+,\times , \div, -']$, which is the same rule shown
    in Figure \ref{fig:trees2}(d).
    \item Non-symmetric $MERGE$ occurs when one object is more deeply nested than the other.
    For instance if $SO_{O1} = [_{operator}\ '+']$ and $SO_{O2} = [_{operator}\ '\div,\times']$ . For these objects Minimal Search chooses the most proximal object as the head. Therefore, $SO_{O1}$ would be used to label the new object as $SO_{new} =[_{operator}\ '+']$, leaving the compliment $SO_{O2}$ inaccessible;
    this is shown in  Figure \ref{fig:trees3}(a).
\end{itemize}

Second, $MERGE$ of an \textit{operator} object $SO_O$ and a \textit{semantic} object $SO_S$,
will produce an \textit{incomplete} object $SO_{new} =  \{SO_{O}, SO_{S} \}$ as shown in Figure \ref{fig:trees2}(a),
but in this case the head of $SO_{new}$ can include a list of operators.
For instance, if $SO_{S} = [_{semantic}\ 'x_1\ ']$ and $SO_O = [_{operator}\ '+,\div']$ the new object
will be labeled as $SO_{new} = [_{incomplete}\ 'x_1 \ +,\div']$, shown in Figure \ref{fig:trees3}(b).

Finally, if $MERGE$ is applied to an \textit{operator} object $SO_{O1}$ and an \textit{incomplete} object $SO_u = \{SP_S, SO_{O2}\}$
the new object  $SO_{new} =  \{SO_{O1}, SO_{u} \}$ will be labeled as an \textit{incomplete} object.
In this case there are two list of operators, those from $SO_{O1}$ and those from $SO_{O2}$.
When the Labeling Algorithm is invoked for $SO_{new}$, the head considers the operators from $SO_{O1}$, not $SO_{O2}$, due to Minimal Search,
since the operator list of the former are closer to the root than the operators from the latter. 
For example, if $SO_{O1} = [_{operator}\ '+,\times']$ and $SO_{u} = [_{incomplete}\ 'x_1 -']$, the new object is labeled as
$SO_{new} = [_{incomplete}\ 'x_1 + \times']$, shown in Figure \ref{fig:trees3}(c).

\subsection{Higher-order $MERGE$ using \textit{semantic} objects}
Some of these cases are presented in Figure \ref{fig:trees4}.
A symmetric $MERGE$ operations between two \textit{semantic} objects follows the same rules described for atomic objects, triggering a Phase Transition and assigning the head for the new object based on the interface condition (depicted in Figure \ref{fig:trees2}(c)).
A $MERGE$  operator between a \textit{semantic} object $SO_{S1}$ and an \textit{incomplete} object $SO_U = \{SO_{S2},SO_O\}$ will also produce a new \textit{semantic} object $SO_{new} = \{SO_{S1},SO_U\}$.
Two possibilities exist depending on the head of $SO_{O}$:
(i) it includes a single operator; or (ii) it includes a list of operators.
Labeling is resolved as follows for each case:
\begin{itemize}
    \item In case (i) $SO_S$ provides the missing term for the incomplete operation encoded in $SO_u$, constructing a computable expression.
    For instance, if $SO_{S1} = [_{semantic} \  'x_1 + x_1\ ']$ and $SO_{U} = [_{incomplete} \  'x_2 + ']$,
    then $SO_{new} = [_{semantic} \  'x_2 + x_1 + x_1\ ']$, as shown in Figure \ref{fig:trees4}(a).

    \item In case (ii), for example with $SO_O = \{'+','-'\}$, all possible combinations
    $'SO_{S1} \ op \  SO_{S2} \ '$ are evaluated $\forall op  \in SO_O$, and the head is assigned based on the interface condition, similar to the case in Figure \ref{fig:trees2}(c).
    For instance if $SO_{S1} = [_{semantic} \  'x_1\ ']$ and $SO_{u} = [_{incomplete} \  'x_2 + \times']$,
    then both $x_2 + x_1$ and $x_2 \times x_1$ are feasible labels for the new object,
    and one is chosen based on the interface condition (randomly or based on an objective function),
    such that the resulting object can either be  $SO_{new} = [_{semantic} \  'x_2 + x_1\ ']$
    or $SO_{new} = [_{semantic} \  'x_2 \times x_1\ ']$; shown in Figure \ref{fig:trees4}(b).
\end{itemize}

\subsection{Higher-order $MERGE$ using \textit{incomplete} objects}
\begin{figure*}
    \centering
    \includegraphics[width=0.95\textwidth]{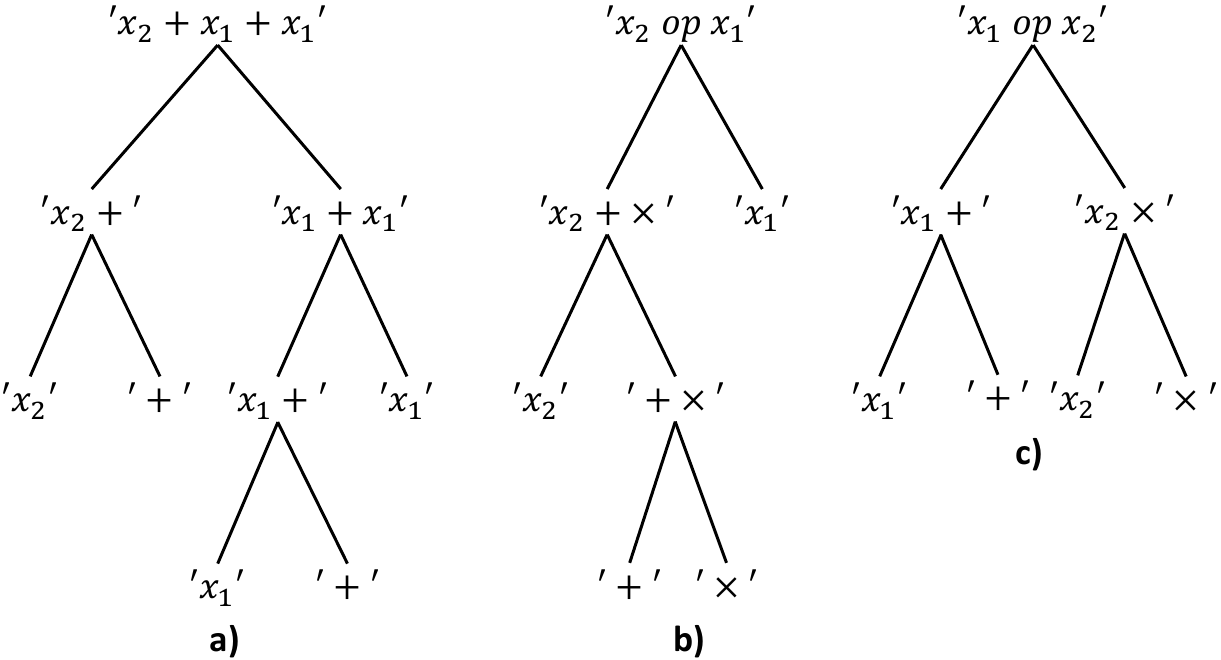}
    \caption{Examples of Higher-Order $MERGE$ with semantic and incomplete objects:
    (a)  an incomplete object $[_{incomplete}\ 'x_2 +']$ and a semantic object
    $[_{semantic}\ 'x_1 + x_1']$ to generate $[_{semantic}\ 'x_2 + x_1 + x_1']$;
    (b)  two incomplete objects $[_{incomplete}\ 'x_1 +']$  and $[_{incomplete}\ 'x_2 \times']$  to generate $[_{semantic}\ 'x_1 \ op \ x_2']$ where $op \in \{ +,\times\}$ that is selected based on the interface condition; and
    (c) two incomplete objects $[_{incomplete}\ 'x_1 +']$ and $[_{incomplete}\ 'x_2 \times']$ to generate $[_{semantic}\ 'x_1 \ op \ x_2']$ where $op \in \{ +,\times\}$ determined as in (b).
    }
    \label{fig:trees4}
\end{figure*}

The final merging condition is when both of the objects $SO_{U1} = \{SO_{S1},SO_{O1}\}$ and $SO_{U2} = \{SO_{S2},SO_{O2}\}$
are \textit{incomplete}, which triggers a Phase Transition, and again several labeling options are available,
resolved using the interface condition.
In this case, all possible combinations
$'SO_{S1} \ op \  SO_{S2} \ '$  $\forall op  \in SO_{O1} \bigcup SO_{O2} $ are evaluated;
an example is depicted in Figure \ref{fig:trees4}(c).

One last comment regarding the operators and operands, since several of the operators used are
not commutative. For this work, we always take the left branch of a newly formed merge tree to
set the first operand. This means that sometimes the best operand order may not be considered.
In future iterations of the proposed approach the operand order may be resolved using
the interface conditions.

\subsection{Nesting of non-linear unary functions}
When the Lexicon includes non-linear unary operators, like $sin()$, $cos()$ or $\sqrt{}$ the following heuristic is used to improve interpretability of the final models.
Consider a $MERGE$ with a \textit{semantic} object $[_{semantic}\ 'sin(x)']$ and an \textit{operator} object $[_{operator}\ 'cos()']$.
Based on the previously outlined rules this would produce $[_{semantic}\ 'cos(sin(x))']$, however nested non-linear operators
are precisely the type of objects that can have a severe negative impact on interpretability \citep{virgolin}.
Therefore, we propose that this $MERGE$ operation instead produces $[_{semantic}\ 'cos(x)']$,
whereby the inner operator $' sin( )'$ is effectively canceled by the top operator $'cos()'$.
This rule was included with the sole purpose of promoting interpretability, since even if such constructions might improve performance,
the models would be black-boxes even if they are expressed symbolically.
This rule can be seen as a relaxation of the Phase Impenetrability Condition, but it is included
in our proposed $MERGE$ operation to promote and enhance interpretability of the constructed objects.

\section{Multi-Derivational Symbolic Regression}
\label{sec:SR}
Building on the $MERGE$ operation defined so far we propose MGP, an MP-based multi-derivational system
for SR tasks. Similar to the relationship between biological evolutionary theory and evolutionary algorithms,
the proposed derivation process is not designed to be a strict simulation of minimalist syntax.
Instead, the core elements and conceptual principles of MP provide inspiration to develop and implement
MGP, while also incorporating necessary enhancements and simplifications that are specific
to the SR task, in order to derive accurate and interpretable models.

\subsection{MP Derivations using $MERGE$}
A complete derivation process is implemented by function $Derive()$ in Algorithm \ref{alg:derivation}.
The key process in the derivation is the $MERGE$ operation defined in the previous section, which
is implemented by way of a wrapper function $Merge()$ (Line 3).
The function incorporates additional hyperparameters that control how the operator
treats high-performance individuals and the interface conditions. 
$Merge()$ also incorporates all of the logic specified in the previous section regarding the labeling process and Phase Transitions.
For simplicity, however, we will refer to
both $MERGE$ and $Merge()$ interchangeably hereafter, unless it is otherwise necessary.

The algorithm begins (Line 1) by initializing the best-so-far syntactic object $SO^*$ with an input object $SO$,
which might be an empty object or an initial baseline object.
The main loop (Line 2-8) iteratively applies $MERGE$ for $n$ steps, and each time a new object $SO_{new}$
is generated it is compared with the current best, potentially replacing it.
Note that $MERGE$ includes four hyperparameters.
First, $p_E$ determines the probability of applying $EM$, which is set to
$p_E = 1$ in this work, leaving the analysis of applying $IM$ for future work.
Second, $e_{sel}$ introduces elitism into the $MERGE$ operation specifying the probability
of choosing the best object in the workspace (lowest RMSE, for instance) as the first object in the merge operation,
while the second object is chosen randomly.
When $e_{sel}=0$ this aligns with MP allowing $MERGE$ to be completely free,
however $e_{sel}$ can be used to bias the derivation.
The third hyperparameter is a binary flag $e_{sur}$ that also introduces elitism into the derivation process.
In this case, when $e_{sur}=0$ the $MERGE$ operation always removes both syntactic objects from the workspace.
On the other hand, when $e_{sur}=1$ if one of the objects used by $MERGE$ is the best object in the workspace this object is not removed from the workspace,
breaking with the canonical MP derivation (Minimum Yield and Resource Restrictions), introducing elitist \textit{survival}.
Finally, $IC$ specifies the interface condition used during Phase Transitions.

Function $ChooseBest()$ selects the best object based on the domain-specific objective function,
such as the root-mean-squared-error (RMSE) or the coefficient of determination ($R^2$) for SR.
The main loop also performs an archive update (Lines 5-7) if flag $a==1$.
Function $UpdateArchive()$ is responsible for determining if $SO_{new}$ will be inserted into archive $\mathcal{A}$,
which is set to a maximum size $n_{\mathcal{A}}$.
The goal of the archive is to store semantically unique
and potentially useful syntactic objects, what are normally referred to as building blocks in evolutionary computation.
The archive is based on novelty search, where solutions are considered novel when they are different
from previously found models \citep{stanley2015greatness}.
To achieve this, $SO_{new}$  is stored in the archive if the following conditions are met:
(1) the archive is not full;
(2) it object does not implement a constant semantic output;
(3) no other object in $\mathcal{A}$ has a semantic vector that is similar to the semantic vector
of the new object $SO_{new}$, determined using the cosine similarity and a threshold $h=0.99$.
The algorithm returns $SO^*$ and $\mathcal{A}$ as outputs.

\begin{algorithm}[t]
\caption{MP Derivation for Symbolic Regression: \newline $  SO^*, \mathcal{A}  \gets Derive(\vec{y}, n, {WS}_0, \mathcal{A}, p_{E},a,e_{sel}, e_{sur},IC,SO) $}
\label{alg:derivation}
\begin{algorithmic}[1]

\Require Target $\vec{y} \in \mathbb{R}^m$, number of steps $n \in \mathbb{N}$, Initial Workspace $WS_0$, Archive $\mathcal{A}$, probability of applying External Merge $p_{E} \in [0,1]$, archive flag $a \in \{false,true\}$, elitist selection probability $e_{sel} \in [0,1]$, elitist survival flag $e_{sur} \in \{0,1\}$, interface condition $IC \in \{Fitness, Random \}$, semantic object $SO$
\Ensure Best semantic object $SO^*$ and updated Archive $\mathcal{A}$

\State Initialize  $SO^* \gets SO$

\For{$i = 1$ to $n$}
    \State $SO_{new}$, $WS_{i}$ $\gets MERGE(WS_{i-1},p_{E}, e_{sel}, e_{sur}, IC)$
    \State $SO^* \gets ChooseBest(SO^*,SO_{new})$ 
    \If{$a== 1$} 
         \State $\mathcal{A} \gets UpdateArchive(\mathcal{A},SO_{new})$
    \EndIf
\EndFor

\State \Return $SO^*,\mathcal{A}$

\end{algorithmic}
\end{algorithm}

\subsection{Iterative MP Derivations}
\label{sec:it}
Given that $MERGE$ can be applied to any syntactic objects in the workspace,
there is a low probability that a single derivation will generate the desired symbolic model.
A more promising approach is to apply this process several times.
Moreover, we can reuse what was constructed during previous derivations, including the best solution found so far $SO^*$ 
and the archive $\mathcal{A}$, a process presented in Algorithm \ref{alg:iterative} for the
$iDerive()$ function, reusing some the hyperparameters from $Derive()$.

\begin{algorithm}[t]
\caption{Iterative Derivations: \newline $ SO^*, \mathcal{A}  \gets iDerive(\vec{y}, n, d, LEX, \vec{cat}, \mathcal{A}_i,\mathcal{A}_o, p_{E},a_o,a_i,b,e_{sel}, e_{sur},IC) $}
\label{alg:iterative}
\begin{algorithmic}[1]

\Require $\vec{y}$, $n$, number of derivations $d \in \mathbb{N}$, lexicon $LEX$ with $p$ categories, vector
specifying number of atoms to generate from each category $\vec{cat}  \in \mathbb{N}^p$, input Archive $\mathcal{A}_i$, output Archive $\mathcal{A}_o$, $p_{E}$, output archive flag $a_o$, input archive flag $a_i \in \{0,1\}$, seed best flag $b \in \{0,1\}$, $e_{sel}$, $e_{sur}$, $IC$
\Ensure Best semantic object $SO^*$ and updated output Archive $\mathcal{A}_o$

\State Initialize  $SO^* \gets \emptyset$

\For{$i = 1$ to $d$}

    \State $WS_0 \gets InitializeWorkspace(LEX,\vec{cat})$
    \If{$a_i== 1$} 
         \State $WS_0 \gets WS_0 \bigcup \mathcal{A}_i$
    \EndIf
    \If{$b== 1$ AND  $i > 1$ } 
         \State $WS_0 \gets WS_0 \bigcup SO^*$
    \EndIf
    \State $SO^*, \mathcal{A}  \gets Derive(\vec{y}, n, {WS}_0, \mathcal{A}_o, p_{E},a_o,e_{sel}, e_{sur},IC,SO^*) $
     \If{$isPerfectFit(SO^*,\vec{y})$} 
         \State \Return $SO^*,\mathcal{A}$
    \EndIf

\EndFor

\State \Return $SO^*,\mathcal{A}$

\end{algorithmic}
\end{algorithm}

First, the best-so-far is initialized in the first line.
The main loop (Lines 2-14) performs $d$ derivations and  several other actions, some of which are activated by control flags.
First, the workspace is initialized by function $InitializeWorkspace()$, that takes the lexicon $LEX$ and a hyperparameter vector $\vec{cat}$ of size $p=6$ in our
experiments; i.e, six operator categories, which are:
\begin{enumerate}
\item linear arithmetic operators ($+$ and $-$);
\item smooth arithmetic operators ($+$, $-$ and $\times$); 
\item arithmetic operators ($+$, $-$, $\times$ and $\div$); 
\item discontinuous arithmetic operator ($\div$); 
\item non-linear trigonometric operators ($cos()$ and $sin()$); and
\item problem variables ($x_1$,$x_2$, etc.).
\end{enumerate}
The values in $\vec{cat}(1)$ to  $\vec{cat}(6)$, respectively specify how many atomic objects to include in the
initial workspace $WS_0$ from each category, randomly choosing between between elements from the same category.
The logic behind these operator categories is based on the potential interpretability of symbolic models built using operators from each category, assuming a subjective scale where:
highly interpretable models only use (1), interpretable models use (2), moderately interpretable models use (3) and models with low interpretabilty also use categories (4) and/or (5). This scale is not rigorous, but it is a useful heuristic that will be used to help drive the derivation process
with a varying focus at different stages of model construction.

After initialization of the workspace (Line 4), hyperparameter $a_i$ determines if the input archive $ \mathcal{A}_i$ is used to seed the initial workspace.
This happens for every derivation, with $ \mathcal{A}_i$ effectively augmenting the initial workspace in a prescribed way.
Similarly (Line 7), flag $b$ is another elitist hyperparamter, adding the best-so-far syntactic object $SO^*$ to the initial
workspace. 
The derivation process from Algorithm \ref{alg:derivation} is then executed, and an early stopping condition is evaluated with
function $isPerfectFit()$ (Line 11), which validates if $SO^*$ perfectly fits the training data, in this work using a tolerance of $0.0001$.

\subsection{Proposed Symbolic Regression Multi-Derivation Algorithm}
\begin{figure*}
    \centering
    \includegraphics[width=0.95\textwidth]{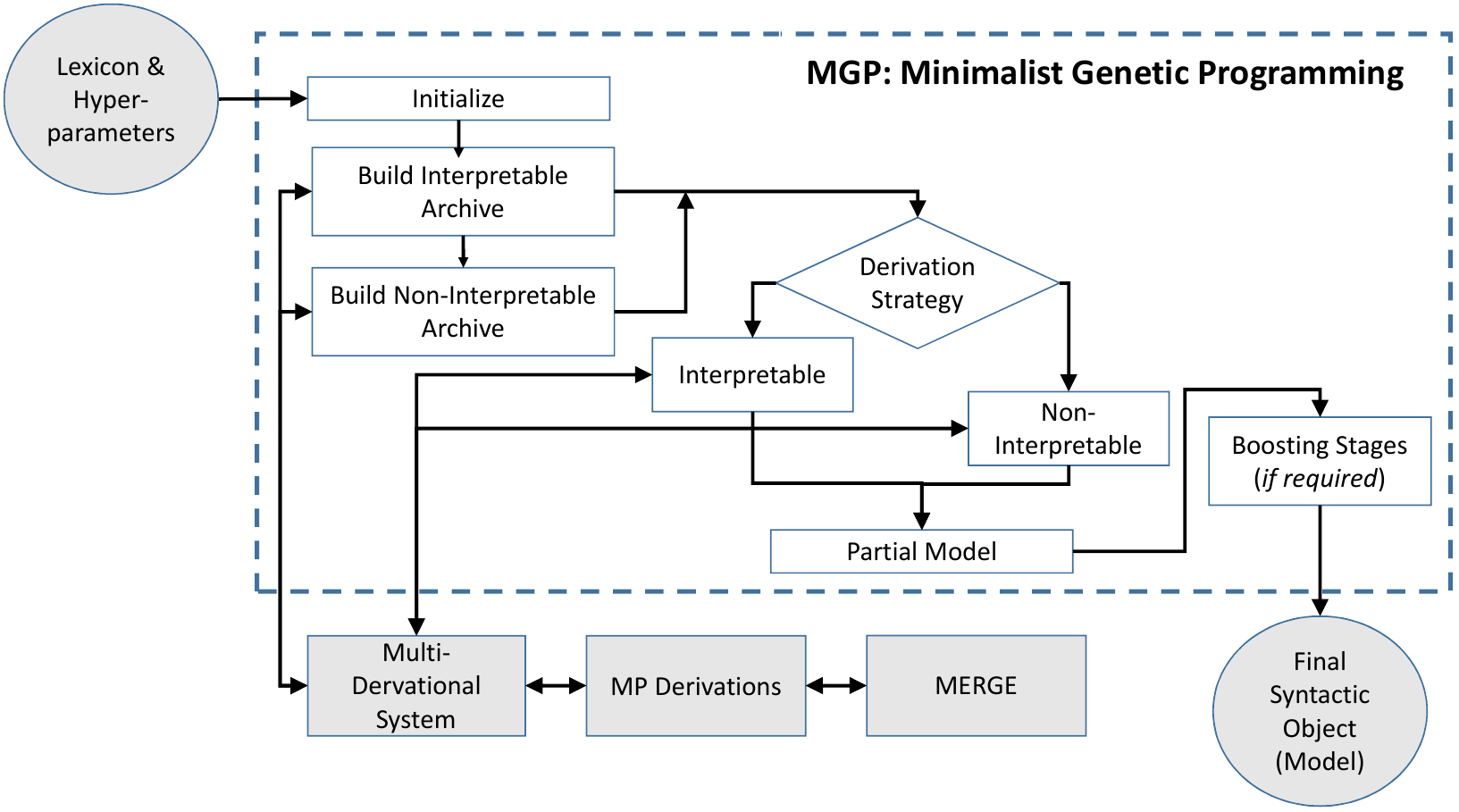}
    \caption{High-level view of Minimalist Genetic Programming.    }
    \label{fig:mgp}
\end{figure*}

Figure \ref{fig:mgp} presents a high level view of the proposed MGP algorithm.
The systems takes as input the training data, a Lexicon, and several hyperparameters.

MGP starts by building two archives of syntactic objects, referred to as an \textit{interpretable}
archive $\mathcal{A}_{I}$ and a \textit{non-interpretable} archive $\mathcal{A}_{NI}$.
While this process could be done to build an arbitrary number of archives, exploring this possibility is left as future work.
The names for these archives are not meant to imply some sort of guarantee for interpretability, or lack there of,
the names are only meant to highlight what types of operators are used and their potential impact on interpretability
based on the informal categories defined before.

The content of these archives are also used to guide the subsequent derivation process.
One path focuses on the use of linear or simple arithmetic operations, while a second potential path is to use more complex, and less interpretable, non-linear functions to construct the final model.

Finally the derivation process allows for up to two boosting stages, to refine the best object found so far.
While further boosting stages are feasible, we limit the number of stages for improved interpretability,
instead of pursuing further, potentially marginal improvements in performance.

A more detailed description of MGP is presented in Algorithm \ref{alg:MGP} that incorporates several calls to $iDerive()$.
Some hyperparameters determine the size of the initial workspace in each derivation ($ops$ and $var$),
while others regulate the intensity of the derivation process and the amount of total generated objects ($n_I$,$n_{NI}$, $d$ and $d_A$).
These hyperprameters are analogous to the population size, number of generations and size limits on trees used in GP.

First, container objects, archives and key hyperparameters are initialized (Line 1).
The derivation begins by using $iDerive()$ to fill archive $\mathcal{A}_{I}$ for a total of $n_I$ steps (Line 2-3),
building a repository of syntactic objects using the simplest types of operators.
The number of atomic elements is defined as $ops$ smooth arithmetic operators and $var$ variables.
Note that the interface condition is set to random ($IC \leftarrow{Random}$),
since the goal of these derivations is not necessarily to find
accurate models, but to build semantically unique objects.
This stage of the process, in other words, is explorative with all flags related to elitism set to $0$ (not used).
Afterward (Line 4), the number and type of lexical items to include as atomic objects is now modified, in order to build more complex and potentially less interpretable objects to be archived in $\mathcal{A}_{NI}$.
In this case using the division operator and trigonometric functions, but also incorporating the previously built archive $\mathcal{A}_{I}$ to seed the workspace (Line 5).

After building both archives the main model building phase is executed, where previously archived objects are used to derive a more expression.
MGP can proceeds in one of two ways, depending on the outcome of the $Compare()$ function that
takes both archives as input and outputs a boolean flag that determines what types of objects to construct (Line 6).
The first option (Lines 8 - 10) is to seed the workspace with the objects in $\mathcal{A}_{I}$ and a reduced number (by a factor $f$) of feature atoms.
These derivations are configured to be more exploitative by setting the interface condition to $Fitness$.
Moreover, exploitation is also increased by seeding the initial workspace with the best-so-far solution ($b=1$),
and never removing the best solution from the workspace ($e_{sur}=1$).
The second option (Lines 12-13) is to use the objects in $\mathcal{A}_{NI}$ to seed the initial workspace instead,
and similarly increase exploitation by inserting and keeping the best solution in the workspace ($b=1$ and $e_{sur}=1$).
However, the interface condition is kept as $Random$, continuing to allow for more exploration during the derivation.
Since the objects in $\mathcal{A}_{NI}$ can express highly non-linear expressions, and are assumed to be
only partial ``building blocks" for a complete model, it is also assumed that the objective function (fitness)
is less predictive of their potential usefulness, unlike the objects in $\mathcal{A}_{I}$.

Before continuing, the heuristic in $Compare()$ has to be explained.
In general, the goal of this function is to determine what type of model should the derivation
process focus on building, based on the information gathered in the archives.
For this work, the criterion is implemented based on the best performance (lowest error)
among all the objects in each archive, referred to as $best_I$ and $best_{NI}$ respectively for $\mathcal{A}_{I}$ and $\mathcal{A}_{NI}$.
Then, if $best_I\times 0.8 \leq best_{NI}$ then $Compare()$ returns $1$ (Line 7),
and a $0$ otherwise (Line 11).
Note that this heuristic may be avoided by performing both derivations (Line 10 and Line 13) and keep the best result,
which can be done in parallel.
However, exploring this option is left as future work.
Before continuing, the the best expression found so far $SO^*$ is simplified
using $Simplify()$ (Line 15),
which performs algebraic simplification and also performs a pruning operation to removing
linear terms that do not contribute to performance.

If a perfect fit has been found (Line 16), the derivation ends and the best object found
is returned.
Otherwise, two potential boosting phased are carried out, by performing a derivation
on the residuals of the best-so-far solution.
The first boosting stage (Line 19-27) is similar to the previous derivation, where emphasis is given
to different archives and operators depending on the result of the $Compare()$ heuristic.
If $Compare()$ returns $1$ the derivation focuses on $\mathcal{A}_{I}$ and restricts the workspace
so simple operators (Line 21-23).
Otherwise (Line 25-26), the derivation incorporates more exploration considering the objects in both
archives, and including trigonometric functions as atomic objects.
However, the boosting phase is in general more elitist, seeding the population with the best-so-far solution
in each derivation, setting $e_{sur} =1$ (never removing the best object from the workspace), and greedily
using the best solution in the workspace with probability $e_{sel}=0.5$.
The bosting expression, and the complete expression, are once again simplified (Line 28).

If the first boosting stage does not produce a perfect solution, a second boosting stage is performed (Lines 29-34).
This stage continues to explore different candidate objects ($IC \leftarrow{Random}$), a different subset of
operators (arithmetic operators, including division), and implements an elitist survival and selection
of the best-so-far object in the workspace.
The final model is once again simplified to improve interpretability before returning it (Line 35).

\begin{algorithm}[t]
\caption{MGP: Minimalist Genetic Programming for Symbolic Regression  \newline $ SO^*  \gets MGP(\vec{y}, n, n_{I}, n_{NI}, d,  d_{A}, ops, var, f, LEX)$}
\label{alg:MGP}
\begin{algorithmic}[1]

\Require $\vec{y}$ , $n$, interpretable archive derivation steps $n_{I} \in \mathbb{N}$, non-interpretable archive derivation steps $n_{I} \in \mathbb{N}$, $d$, archiving derivations $d_A \in \mathbb{N}$, number of atomic operators $ops \in \mathbb{N}$, 
number of atomic variables $var \in \mathbb{N}$. reduction factor $f \in [0,1]$,
lexicon $LEX$ 
\Ensure Best semantic object $SO^*$

\State  $SO^* \gets \emptyset$, $SO_{B1} \gets \emptyset$, $SO_{B2} \gets \emptyset$, $\mathcal{A}_{I} \gets \emptyset$, $\mathcal{A}_{NI} \gets \emptyset$, $p_E \gets 1$, $IC \gets Random$

\State $\vec{cat}_{t} \gets [0,ops,0,0,0,var]$ \Comment{\textbf{Build Interpretable Archive}}
\State $ SO^*, \mathcal{A}_I  \gets iDerive(\vec{y}, n_I, d_A, LEX, \vec{cat}_{t}, \emptyset, \mathcal{A}_I, p_{E},1,0,0,0,IC)$

\State $\vec{cat}_{t} \gets [0,0,0,ops,ops,var\times f]$ \Comment{\textbf{Build Non-Interpretable Archive}}

\State $ SO^*, \mathcal{A}_{NI}  \gets iDerive(\vec{y}, n_{NI}, d_A, LEX, \vec{cat}_{t}, \mathcal{A}_I, \mathcal{A}_{NI}, p_{E},1,1,0,0,IC)$

\State bool  $Interpret \gets Compare(\mathcal{A}_I,\mathcal{A}_{NI})$
\If{$Interpret==1$}  \Comment{\textbf{Interpretable or Non-Interpretable Derivation}}
  \State $\vec{cat}_{t} \gets [0,ops,0,0,0,var \times f]$
  \State $IC \gets Fitness$
  \State $ SO^* \gets iDerive(\vec{y}, n_I, d, LEX, \vec{cat}_{t}, \mathcal{A}_I, \emptyset, p_{E},0,1,1,0,1,IC)$      
\Else
\State $\vec{cat}_{t} \gets [0,ops,0,0,0,0]$
\State $ SO^* \gets iDerive(\vec{y}, n_I, d, LEX, \vec{cat}_{t}, \mathcal{A}_{NI}, \emptyset, p_{E},0,1,1,0,1,IC)$
\EndIf

\State $SO^* \gets Simplify(SO^*)$ \Comment{\textbf{Simplify the expression}}

\If{$isPerfectFit(SO^*,\vec{res}) == true$}
    \State \Return $SO^*$ \Comment{\textbf{Return if Solution is Perfect}}
\EndIf

    \State $\vec{res} \gets Residuals(SO^*,\vec{y})$ \Comment{\textbf{Begin First Boosting Stage}}  
         \If{$Interpret==1$}
            \State $IC \gets Fitness$
            \State $\vec{cat}_{t} \gets [0,ops,0,0,0,var]$
            \State $ SO_{B1}  \gets iDerive(\vec{res}, n, d, LEX, \vec{cat}_{t}, \mathcal{A}_{I}, \emptyset,
            p_{E},0,1,1,0.5,1,IC)$         
        \Else
            \State $\vec{cat}_{t} \gets [0,ops,0,0,ops \times f,var]$
            \State $ SO_{B1} \gets iDerive(\vec{res}, n, d, LEX, \vec{cat}_{t}, \mathcal{A}_{NI} \cup  \mathcal{A}_{I}, \emptyset,
            p_{E},0,1,1,0.5,1,IC)$
        \EndIf

\State $SO^* \gets Simplify(SO^* + Simplify(SO_{B1}))$ \Comment{\textbf{Simplify the expression}}

\If{$isPerfectFit(SO^*,\vec{res}) == false$}
        
        \State $\vec{res} \gets Residuals(SO^*+SO_{B1},\vec{y})$ \Comment{\textbf{Second Boosting if Solution Not Perfect}}
        \State $IC \gets Random$
        \State $\vec{cat}_{t} \gets[0,0,ops,0,0,var]$
        \State $ SO_{B2}  \gets iDerive(\vec{res}, n, d, LEX, \vec{cat}_t, \emptyset, \emptyset, p_{E},0,0,1,0.5,1,IC)$

\EndIf

\State \Return $Simplify(SO^* + Simplify(SO_{B2}))$

\end{algorithmic}
\end{algorithm}

\subsubsection{Model Coefficients and Computational Complexity}
The only terminal elements considered in MGP are problem features or variables, omitting constants or other types of coefficients.
While in traditional GP numerical constants are often included as part of the terminal set,
search operators in GP are ill-equipped to optimize them, since they focus on structural modifications.
In fact, bloat may be partially caused by the inability of GP search operators to properly fit simple models \citep{Trujillo2018}.
For this reason, most state-of-the-art SR methods integrate the structural exploration performed by GP with a numerical optimizer
that fits model parameters \citep{srbench2}.
We assume that a similar principle holds for $MERGE$, and have therefore decided to omit constants and model coefficients from the lexicon.
However, it would be simple to include a fine-tuning process for all computable objects in MGP using a numerical optimizer,
without modifying the rest of the derivation process, potentially implemented as part of the Phase Transition in Figure \ref{fig:mp}.
This would allow $MERGE$ to focus on building syntactic structure, while the optimizer would fit all implicit parameter coefficients.
This possibility is left for future work, with the experimental work presented in the following section focusing on tasks
where deriving model structure is the main goal.

Finally, the computational complexity of MGP is $O(mnd)$, where $m$ is the number of fitness
cases, $n$ is the number of derivation steps and $d$ the number of derivations.
Given its sequential nature, however, the potential for improved efficiency using parallel
processing depends on the amount of feedback between derivations in $iDerive()$.
If no archiving is performed, and if the initial workspace is not seeded with the
best so far object, then it is trivial to parallelize all the calls to $Derive()$.

\section{Experimental Analysis}\label{sec:experiments}
Experimental work focuses on synthetic SR benchmarks for two main reasons.
First, they have a ground truth symbolic expression, used evaluate the ability of MGP to recover the true symbolic models.
However, we also include symbolic models that are not constructable with the Lexicon used, to evaluate the ability of MGP as an approximation
method.
Second, the synthetic benchmarks can be chosen such that the role of model coefficients or parameters is practically eliminated, focusing the
evaluation on the ability of MGP to recover the underlying structure of the ground truth equation, and not the capacity of he system to perform parameter fitting\footnote{Source code will be made available once the paper is accepted for publication}.

MGP is evaluated from three perspectives.
First, the ability to recover the ground truth model.
Second, the typical train and test set performance evaluation.
For testing in particular, focus is on the ability of the models to extrapolate to out of distribution samples.
Finally, we evaluate some of the internal dynamics of MGP, including the workspace composition,
convergence curves and the relevance of boosting on different problems.

\subsection{Experimental Setup}
The chosen problems are from \citep{Uy2010}, which have been used extensively to benchmark GP systems.
While the expressions are not particularly complex, standard GP will struggle to recover the true ground truth model
in most runs.
The problems are summarized in Table \ref{tab:problems}, including the training domains.
For all problems $100$ equidistant points in the training domain are used to evaluate the objective function
used to guide the MGP derivations.
Additionally, based on the Lexicon used by MGP (see Section \ref{sec:it}), some of these problems
are solvable or unsolvable, in the sense that the ground truth model can be expressed using
the a subset of the lexical items.
The unsolvable problems are used to evaluate the ability of MGP to find approximate solutions.

Problem Nguyen-7 does not include the constants originally included in \citep{Uy2010}, since MGP is not yet
configured to search for optimal coefficients.
For the group of \textit{Solvable} problems, MGP is evaluated based on its ability to recover the ground truth model.
On the other hand, for the \textit{unsolvable}  problems MGP is evaluated based on the ability to fit and generalize with approximate models.
MGP is also compared with a naive model approximation using a polynomial approximation.

\begin{table}[t]
    \centering
    \scriptsize
    \caption{Benchmark SR problems. Problem Nguyen-7 does not include the constants originally included in \citep{Uy2010}. For 2D problems (Nguyen-9, Nguyen-10, Nguyen-11 and Nguyen-12), the training domain is interpreted     as a 2D mesh, with the same range for both dimensions.}

	\begin{tabular}{lccc}
    \toprule
	{\bf Problem}   	& {\bf Equation} & {\bf Training Domain} &  {\bf Experiment}  \\
    
    \midrule
        Nguyen-1      & $x_1 + x_1^2 + x_1^3$    &   $[-5,5]$     &  Solvable  \\
        Nguyen-2      & $x_1 + x_1^2 + x_1^3 + x_1^4$    &   $[-5,5]$     &  Solvable  \\
        Nguyen-3      & $x_1 + x_1^2 + x_1^3 + x_1^4 + x_1^5$    &   $[-5,5]$     &  Solvable  \\
        Nguyen-4      & $x_1 + x_1^2 + x_1^3+ x_1^4 + x_1^5 + x_1^6$    &   $[-5,5]$     &  Solvable  \\
        Nguyen-5      & $sin(x_1^2)\times cos(x_1)$    &   $[-5,5]$     &  Solvable  \\
        Nguyen-6      & $sin(x_1) + sin(x_1+x_1^2)$    &   $[-5,5]$     &  Solvable  \\
        Nguyen-7      & $log(x_1+1) + log(x_1^2)$    &   $(0,5]$     &  Unsolvable  \\
        Nguyen-8      & $\sqrt{x_1}$    &   $(0,5]$     &  Unsolvable  \\
        Nguyen-9      & $sin(x_1) + sin(x_2^2)$    &   $[-5,5]$     &  Solvable  \\
        Nguyen-10     & $sin(x1)\times cos(x2)$    &   $[-5,5]$     &  Solvable  \\
        Nguyen-11     & $x_1^{x_2}$    &   $(0,5]$     &  Unsolvable  \\
        Nguyen-12     & $x_1^4 - x_1^3 + x_2^2 - x_2$    &   $[-5,5]$     &  Solvable  \\

	\bottomrule
	\end{tabular}

    \label{tab:problems}
\end{table}

The main MGP hyperparameters are given in Table \ref{tab:params}.
A deeper analysis of algorithm hyperparameters is left for future work.
For now, an informal exploration of hyperparameter values was performed, considering $\pm 25\%$  of the values used in this work.
While most configurations achieved similar performance, the values used in the following experiments achieved the best results.
For each problem, 30 independent runs are executed.

\begin{table}[t]
    \begin{center}
    \footnotesize
    \caption{MGP hyperparameters.}
\label{tab:params}
\begin{tabular}{lcc}
\hline
{\bf Hyperparameter}   	& {\bf Symbol} & {\bf Value} \\
\hline
Derivation Steps    & $n$  &  100\\
\hline
Interpretable archive steps     &  $n_I$  &  75  \\
\hline
Non-interpretable archive steps     &  $n_{NI}$  & 35 \\
\hline
Interpretable archive size     &  $|\mathcal{A}_I|$  &  35 \\
\hline
Non-interpretable archive size     &  $|\mathcal{A}_I|$  & 45  \\
\hline
Archiving derivations    &  $d_A$  &   100 \\
\hline
Total Derivations   &  $d$  &  1000 \\
\hline
Atomic operators  &  $ops$  &   20\\
\hline
Atomic variables  &  $var$  &   50 \\
\hline
Reduction factor  &  $f$  &   0.25\\
\hline
Fitness Interface Condition &  $IC$  &   RMSE\\
\hline

\end{tabular}
    \end{center}

\end{table}

\subsection{Results}
This section presents MGP performance on each of the benchmark problems, presenting slightly different analysis
for the Solvable and Unsolvable problems.

\subsubsection{Results on Solvable Benchmarks}

Table \ref{tab:rate} presents the number of percentage of runs that generated the groundhog truth equation.
All problems are solved in over $93\%$ of the runs, with a $100 \%$ solution rate in all but three problems.
In all cases we consider mathematically equivalent models to be the same.
For instance, problem Nguyen-6 has two potential solutions, since
$sin(x_1) + sin(x_1+x_1^2) = sin(x1) + cos(x1.^2).*sin(x1) + sin(x1.^2).*cos(x1)$.
Both represent perfect solutions, since they mathematically equivalent.
In general, MGP achieves close to perfect behavior on these benchmarks, with only a few exceptions.
Table \ref{tab:rate} table also presents the percentage of runs that required each of the boosting stages.

\begin{figure}[t]
	\centering
	\subfigure[Nguyen-1]{\includegraphics[width=.32\textwidth]{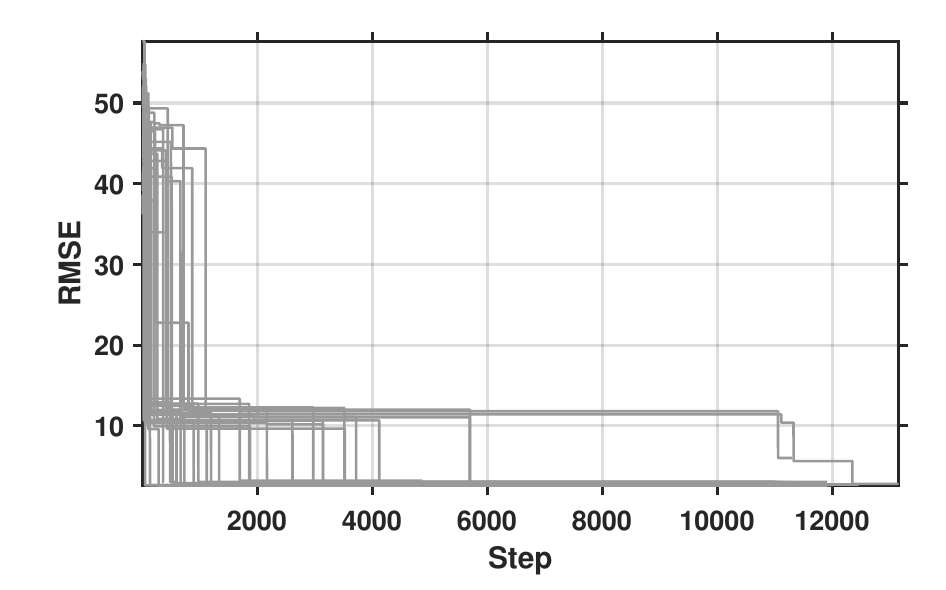}}
    \subfigure[Nguyen-2]{\includegraphics[width=.32\textwidth]{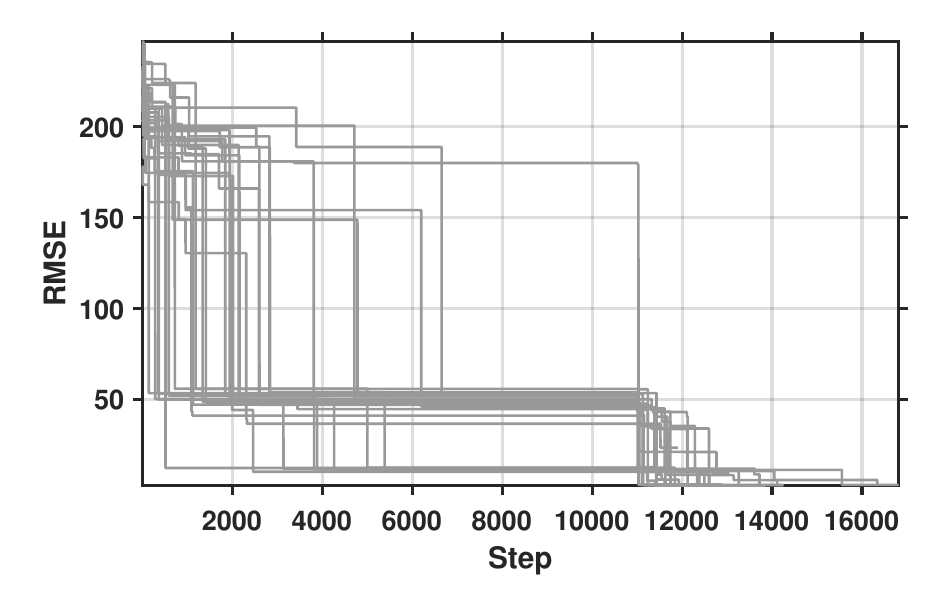}}
    \subfigure[Nguyen-3]{\includegraphics[width=.32\textwidth]{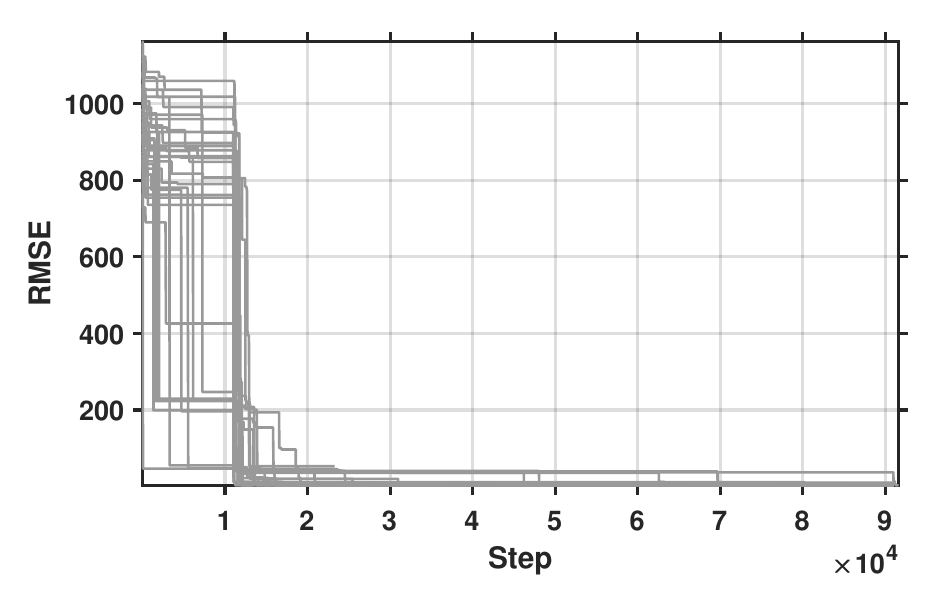}}

    \subfigure[Nguyen-4]{\includegraphics[width=.32\textwidth]{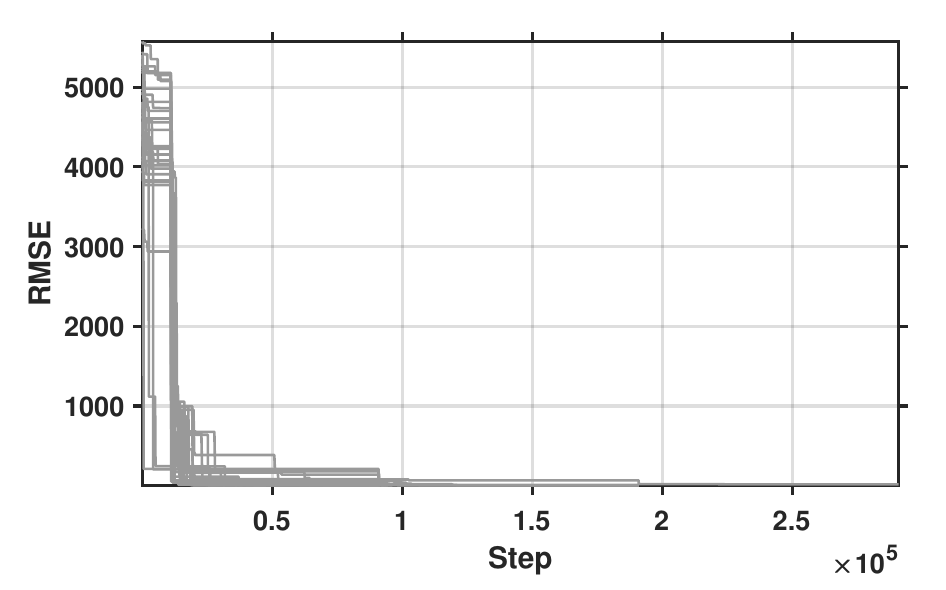}}
    \subfigure[Nguyen-5]{\includegraphics[width=.32\textwidth]{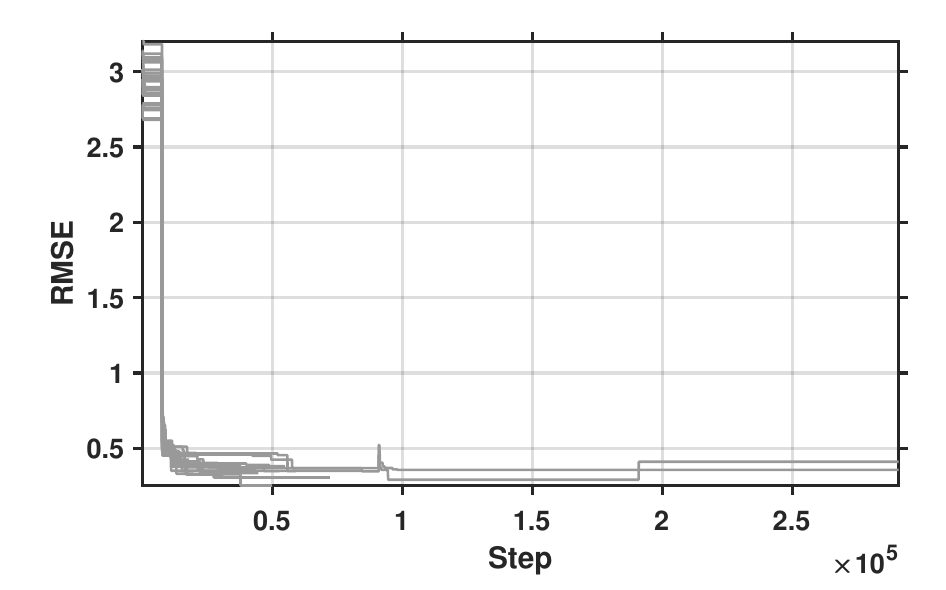}}
    \subfigure[Nguyen-6]{\includegraphics[width=.32\textwidth]{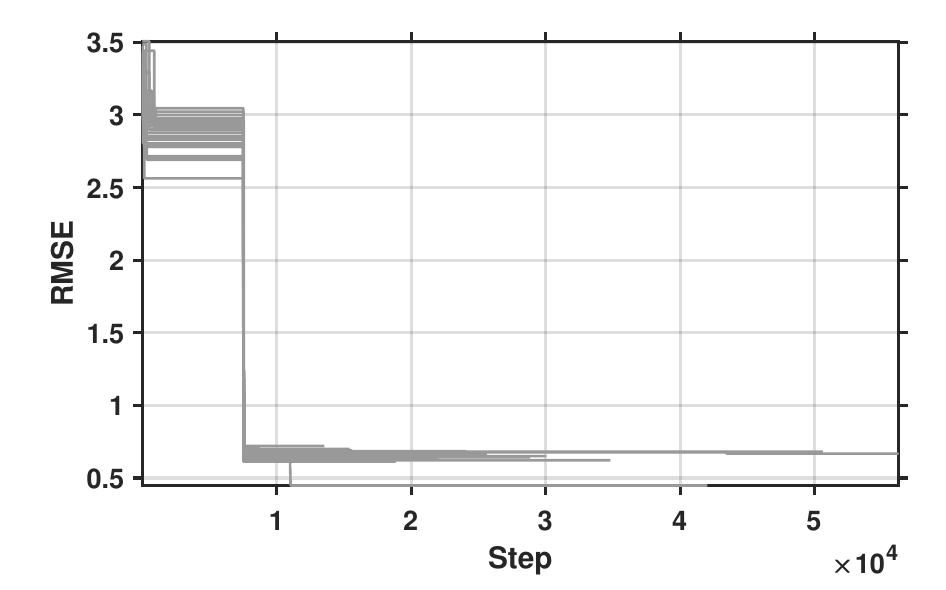}}

    \subfigure[Nguyen-9]{\includegraphics[width=.32\textwidth]{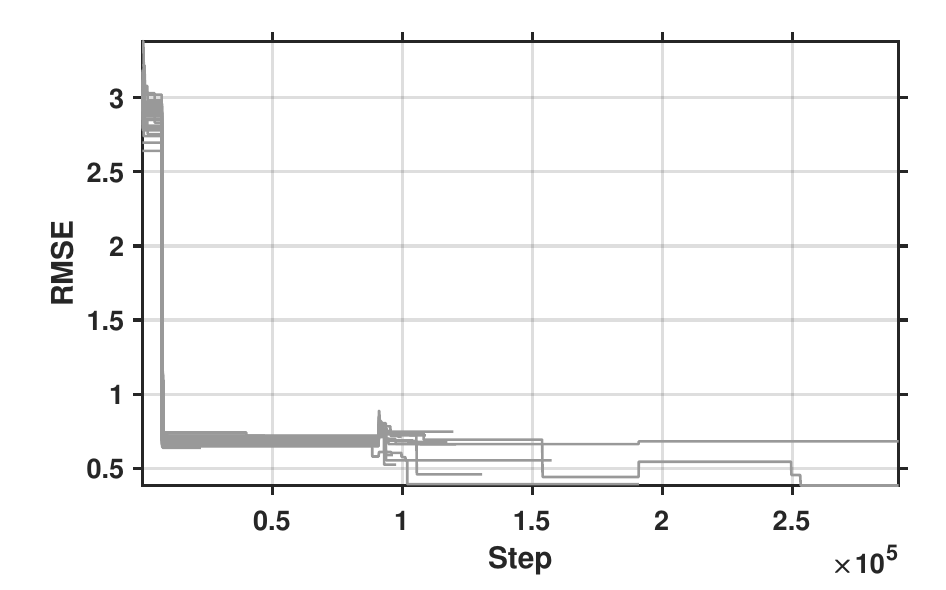}}
    \subfigure[Nguyen-10]{\includegraphics[width=.32\textwidth]{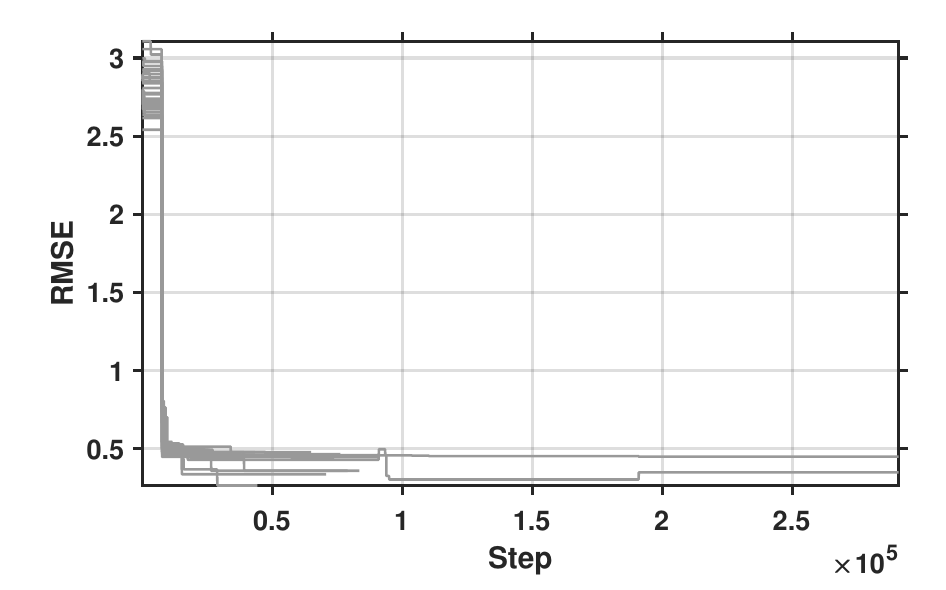}}
    \subfigure[Nguyen-12]{\includegraphics[width=.32\textwidth]{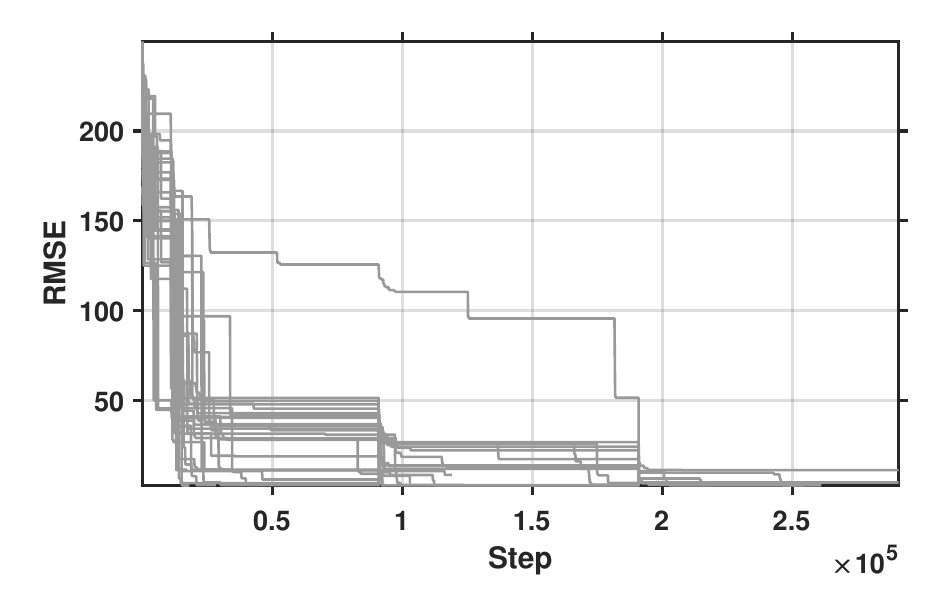}}
	\caption{Convergence plots of MGP on each of the Solvable benchmark problems, showing one curve
    for each run.} 
	\label{fig:convergence}
\end{figure}

First, Figure \ref{fig:convergence} presents the convergence behavior of the algorithm.
For each problem, the RMSE of the best so far model is plotted over all derivation steps, for each of the 30 runs.
Since the algorithm stops once a perfect fit is found, different runs require a different amount of total
derivation steps.
The curves are not monotonic, since the model can decrease in performance (increase in RMSE) during the boosting phases
since the initial  boosting expressions found my be far from optimal.
These plots consider all the steps taken before a perfect model is found, with curves covering the entire
domain indicating that a perfect model was never produced.

Most plots are similar, and reminiscent to the types of convergence plots produced in GP runs.
With a series of large discrete steps towards optimal performance.
For some problems, the horizontal axis is larger, since it accounts for runs that required boosting derivations.
It is possible to see clear consistent convergence patterns in all problems, with only a few outlier cases.

\begin{figure}[t]
	\centering
	\subfigure[Nguyen-1]{\includegraphics[width=.32\textwidth]{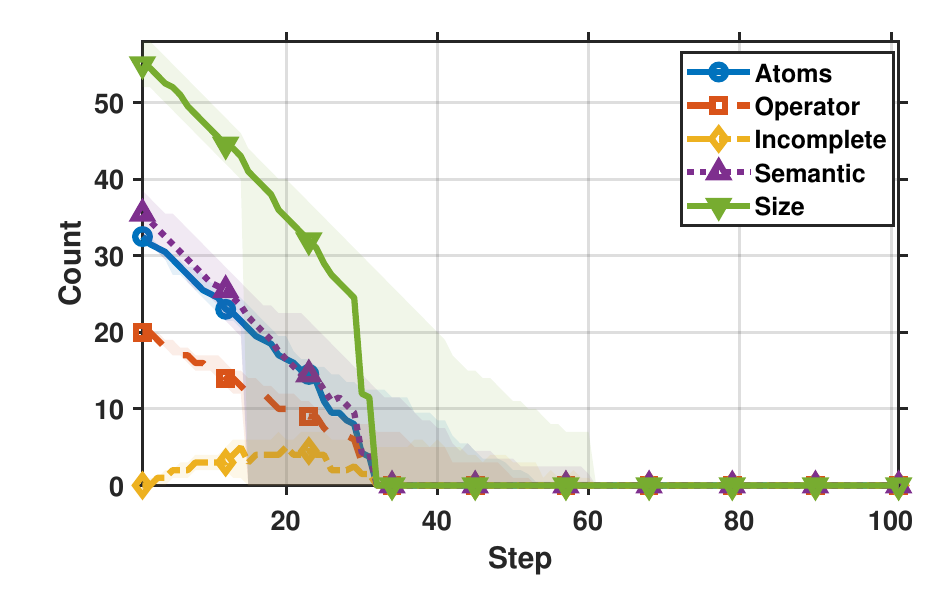}}
    \subfigure[Nguyen-2]{\includegraphics[width=.32\textwidth]{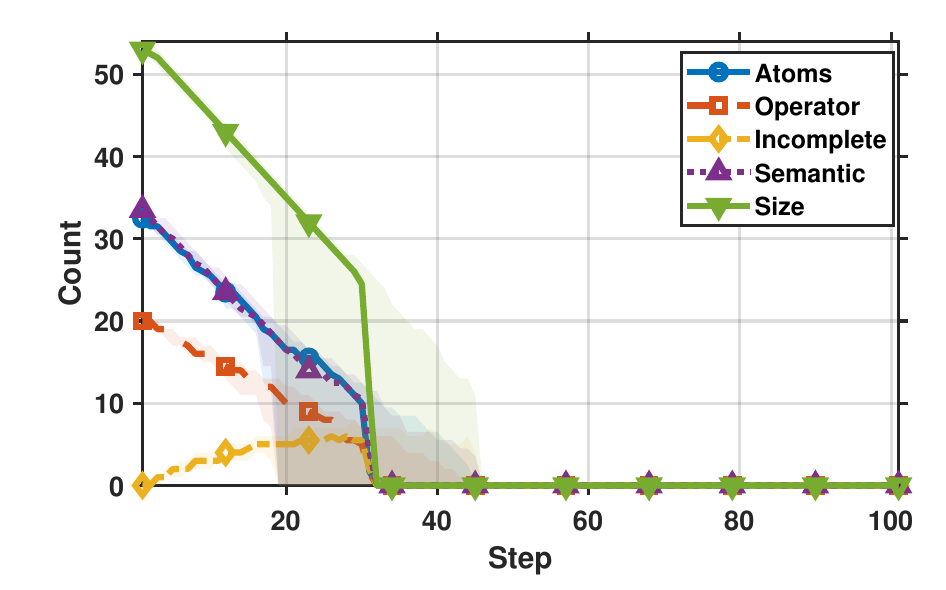}}
    \subfigure[Nguyen-3]{\includegraphics[width=.32\textwidth]{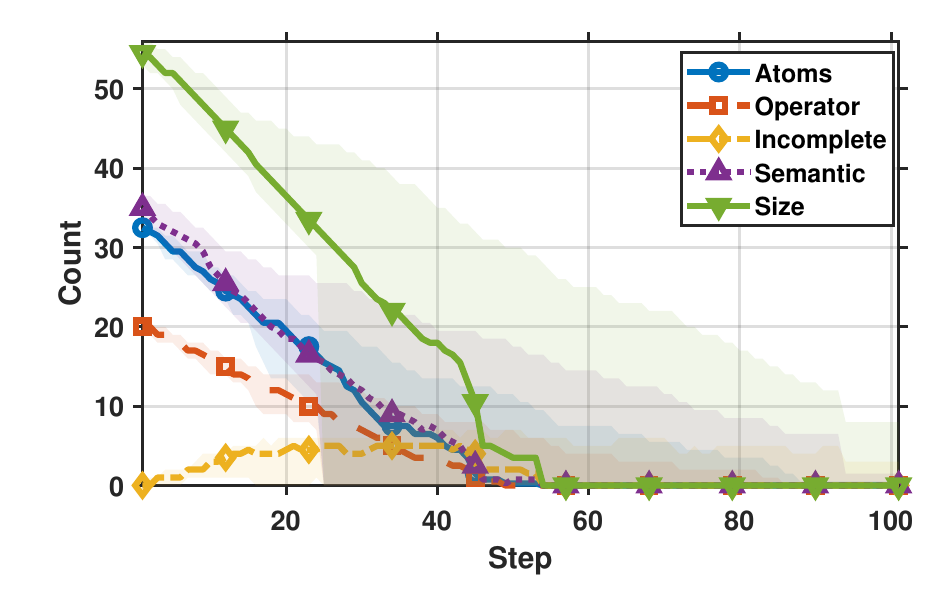}}

    \subfigure[Nguyen-4]{\includegraphics[width=.32\textwidth]{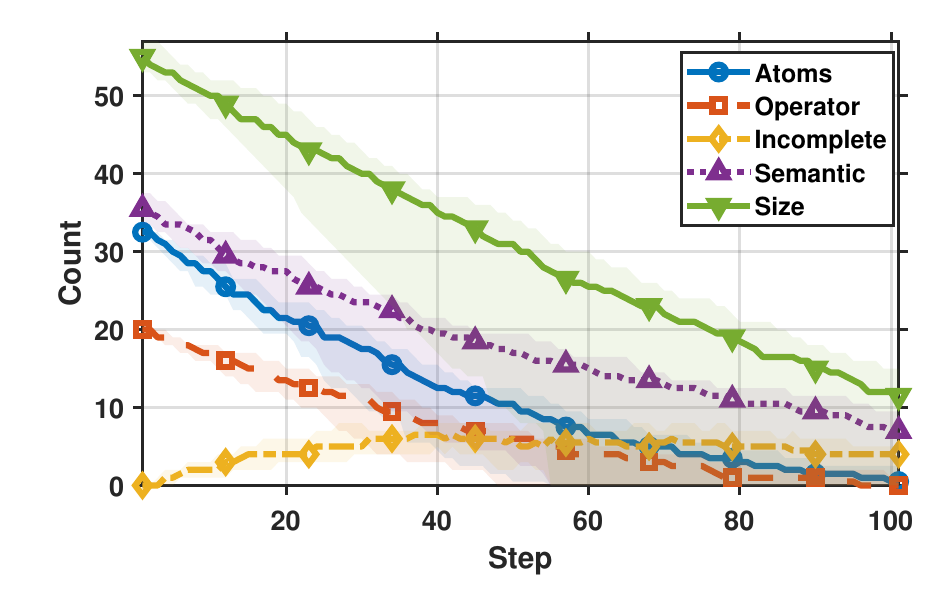}}
    \subfigure[Nguyen-5]{\includegraphics[width=.32\textwidth]{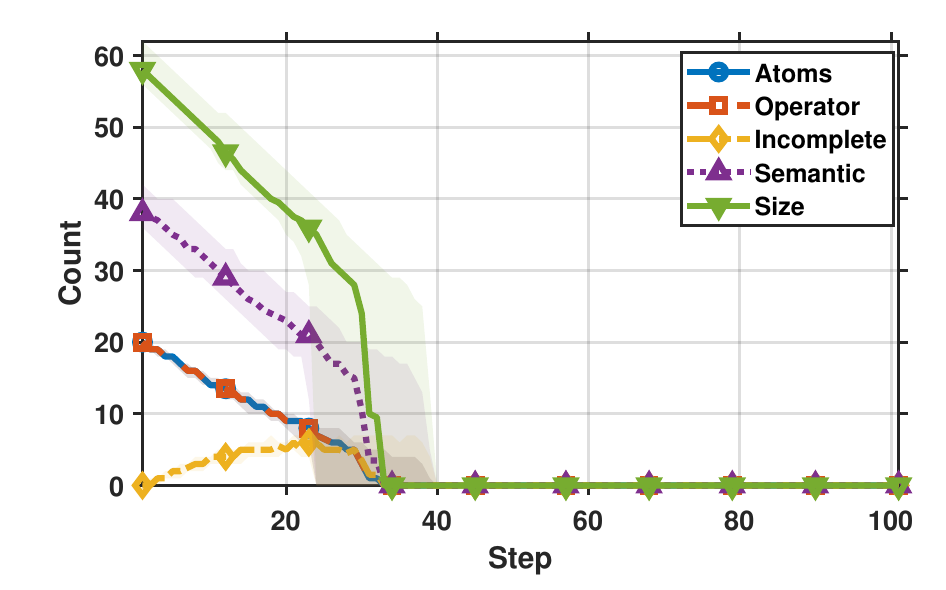}}
    \subfigure[Nguyen-6]{\includegraphics[width=.32\textwidth]{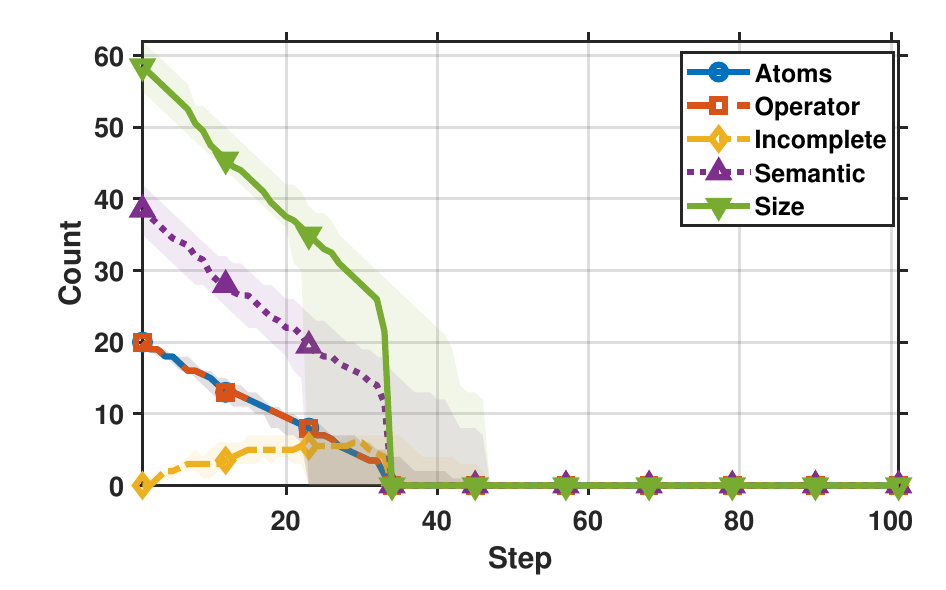}}

    \subfigure[Nguyen-9]{\includegraphics[width=.32\textwidth]{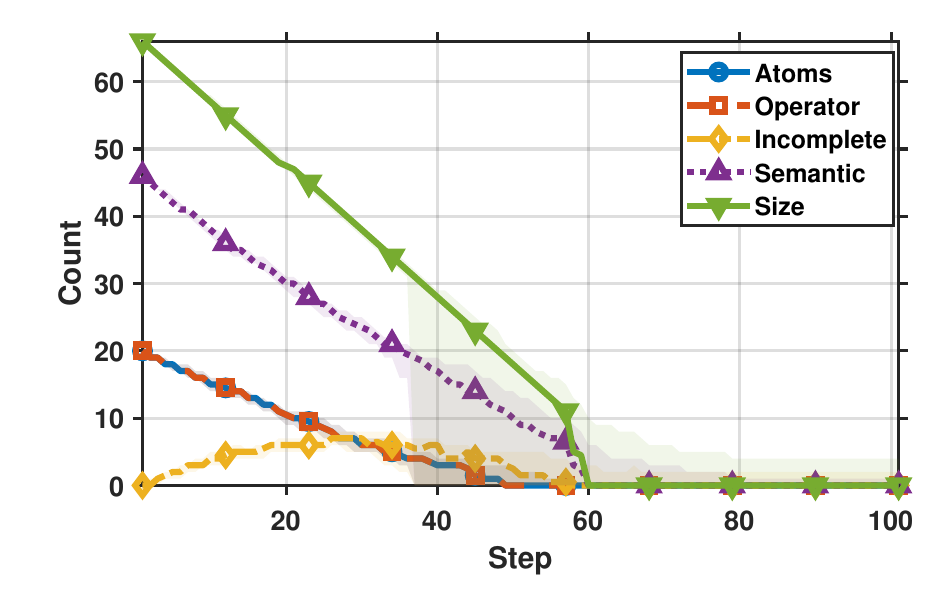}}
    \subfigure[Nguyen-10]{\includegraphics[width=.32\textwidth]{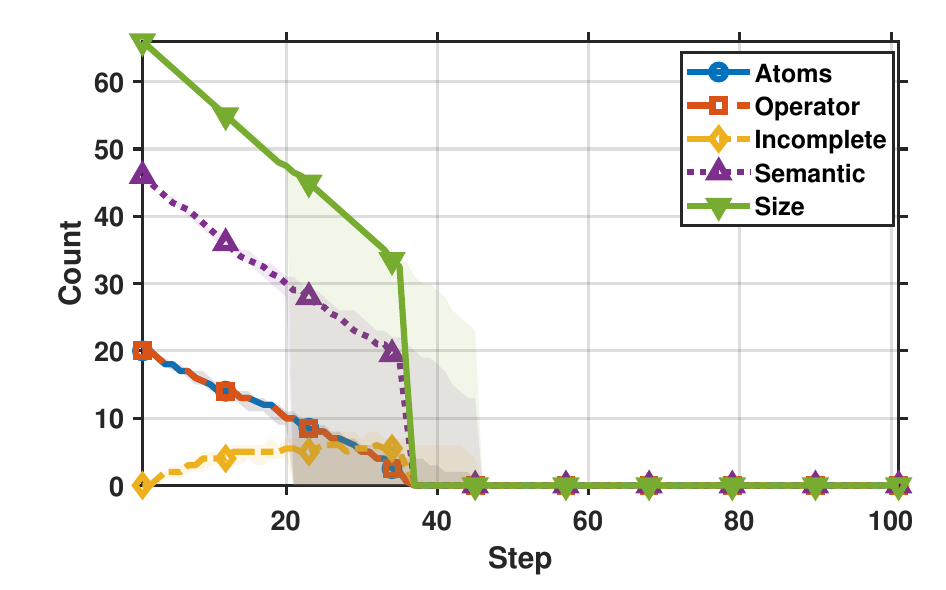}}
    \subfigure[Nguyen-12]{\includegraphics[width=.32\textwidth]{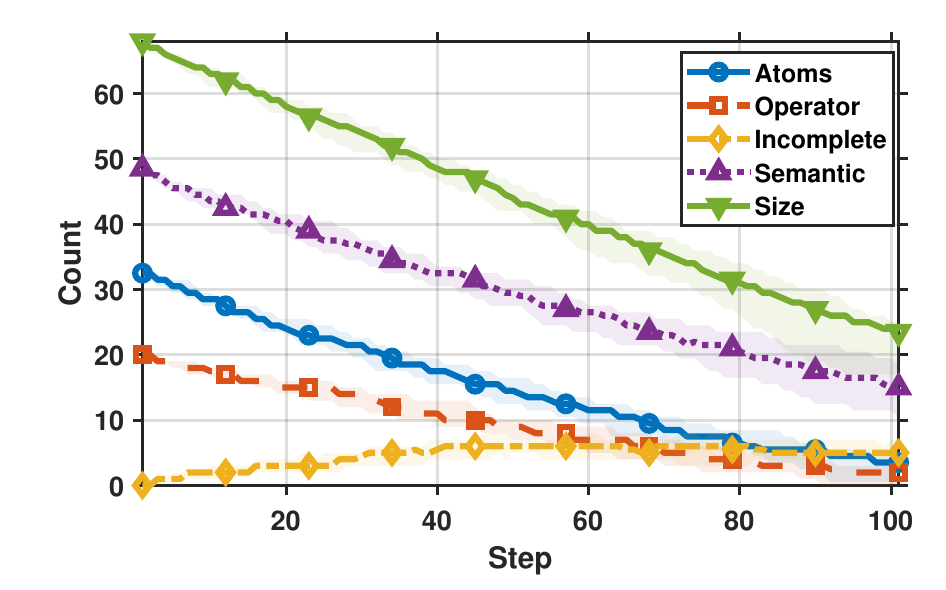}}
	\caption{Workspace composition during the final main derivation for Solvable benchmarks, showing the number of:
    total objects, and atomic, operator, incomplete and semantic objects.} 
	\label{fig:ws}
\end{figure}

The composition of the workspace is explored in Figure \ref{fig:ws}, focusing on how the workspace is composed during the final derivation of the complete model before boosting (Line 10 or Line 13 in Algorithm \ref{alg:MGP}).
The curves show the median number of syntactic objects in the workspace, with the inter-quartile range as a shaded band.
Curves are shown for the total number of syntactic objects (workspace size), as well as the number
of atomic elements, operators, incomplete and semantic objects.
Note the workspace will include some atomic elements, as well as objects from one
of the two archives.

Most of these plots show similar trends, with Nguyen-9 deviating slightly, and Nguyen-4 and Nguyen-12
deviating sharply.
Note that these problems (see Table \ref{tab:rate}) also have the largest proportion of runs
that required at least one boosting phase.
In other words, the final derivation of MGP in these problems did not yet find perfect model.
On the other hand, for all other problems, the final derivation also represents the derivation where
the perfect solution was constructed, hence the truncated curves.
In all cases, the number of objects decreases as objects are merged, with incomplete objects being the only ones
that increase in frequency over time at the beginning of the derivations.
For the problems where an optimal solution is not found, the number of incomplete objects never decreases, and remains
constant after that initial increase.

\begin{table}[t]
    \centering
    \scriptsize
    \caption{MGP performance on Solvable benchmarks, showing the number of times a run produced
    the correct model, from a total of 30, as well as the percentage of runs that required each of the
    boosting stages.}

	\begin{tabular}{lcccc}
    \toprule
	{\bf Problem}   	&  {\bf Successful Runs} & {\bf Solution Percentage} & {\bf 1st Boosting} & {\bf 2nd Boosting} \\
    
    \midrule
        Nguyen-1       & 30   &   $100 \%$  &   $0 \%$  &   $0 \%$   \\
        Nguyen-2       & 30   &   $100 \%$  &   $0 \%$  &   $0 \%$   \\
        Nguyen-3       & 30   &   $100 \%$  &   $30 \%$  &   $0 \%$   \\
        Nguyen-4       & 30   &   $100 \%$  &   $73 \%$  &   $16 \%$   \\
        Nguyen-5       & 28   &   $93.3 \%$  &   $6 \%$  &   $6 \%$   \\
        Nguyen-6       & 30   &   $100 \%$  &   $0 \%$  &   $0 \%$   \\
        Nguyen-9       & 28   &   $93.3 \%$  &   $46 \%$  &   $6 \%$  \\
        Nguyen-10      & 28   &   $93.3 \%$  &   $6 \%$  &   $6 \%$   \\
        Nguyen-12      & 30   &   $100 \%$  &   $83 \%$  &   $40 \%$   \\

	\bottomrule
	\end{tabular}

    \label{tab:rate}
\end{table}

Over all runs, and over all problems, the ground truth model was found $97 \%$ of the time, with only six runs
producing the wrong model (out of 270). However, these cases are interesting to analyze.
For Nguyen-5, the first incorrect model was $- sin(x_1^2) \times cos(x_1)^2$, which includes most of the correct structure.
On the other hand, the other case is a very large expression, basically a product of over a dozen trigonometric functions.

On Nguyen-9, both of the incorrect models identified one part of the
solution, and can be expressed as $sin(x+1) + incorrectTerm$.
Partially recovering an element of the ground truth expression.
Finally, on Nguyen-10, both of the incorrect solutions
where very different from the ground truth, overly relying
on products of many trigonometric functions.
These results suggest that, when the derivation fails, it can still
recover some elements of the desired expression, while completely failing
in other cases.

\subsubsection{Results on Unsolvable Benchmarks}
Performance on the unsolvable benchmarks is presented next.
As stated before, given the Lexicon defined for MGP, it is impossible for the algorithm to derive the ground truth model.
Similarly for GP, if the primitive set does not include the necessary operators, the best that can be produced
is an approximate model.

\begin{figure}[t]
	\centering
	\subfigure[Nguyen-7]{\includegraphics[width=.32\textwidth]{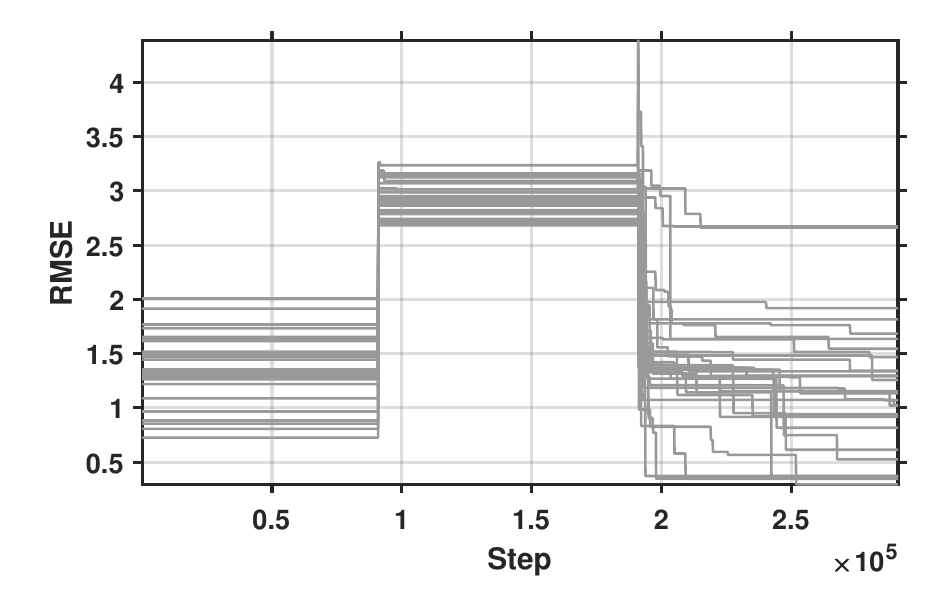}}
    \subfigure[Nguyen-8]{\includegraphics[width=.32\textwidth]{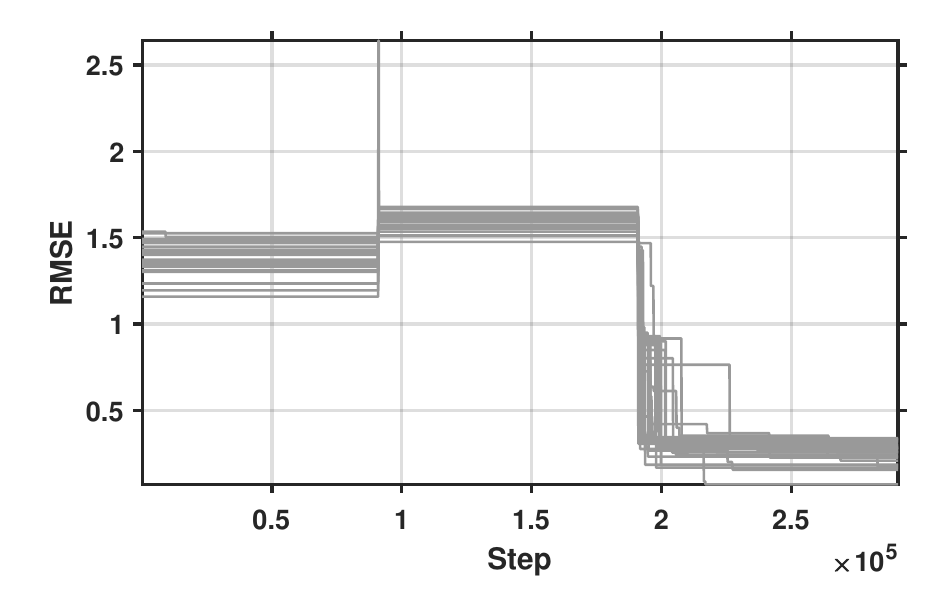}}
    \subfigure[Nguyen-11]{\includegraphics[width=.32\textwidth]{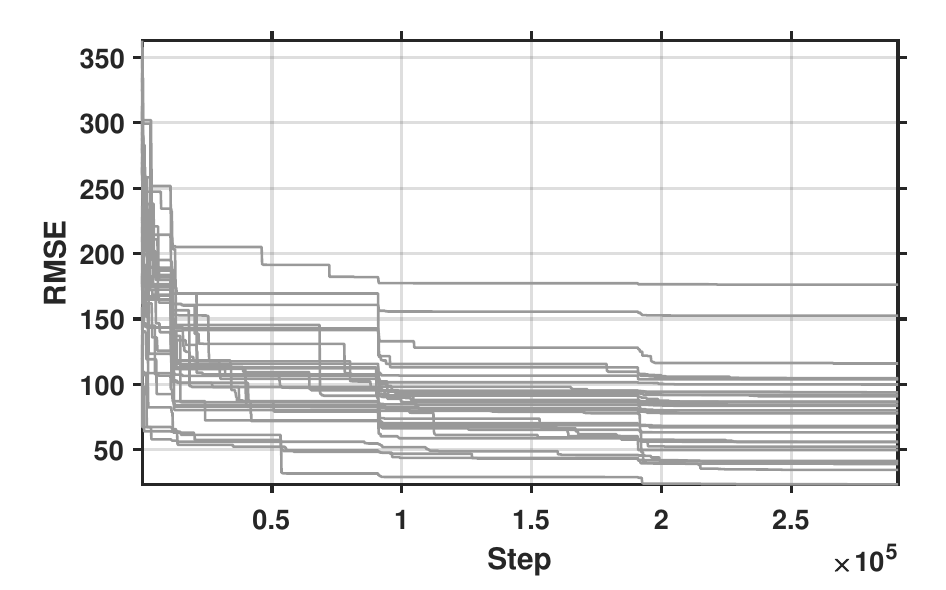}}

	\caption{Convergence plots of MGP on each of the unsolvable benchmark problems, showing one curve
    for each run.} 
	\label{fig:convergence2}
\end{figure}

\begin{figure}[t]
	\centering
	\subfigure[Nguyen-7]{\includegraphics[width=.32\textwidth]{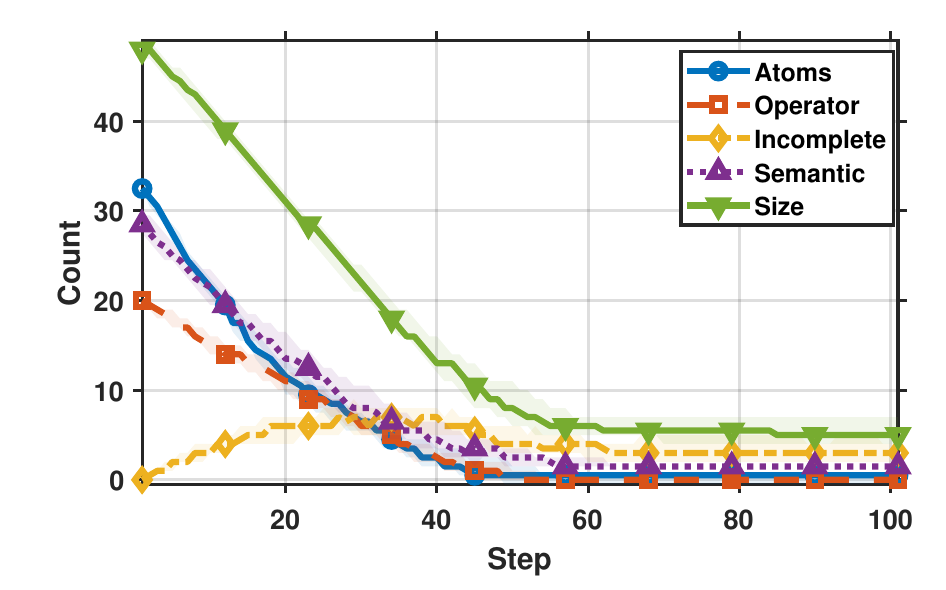}}
    \subfigure[Nguyen-8]{\includegraphics[width=.32\textwidth]{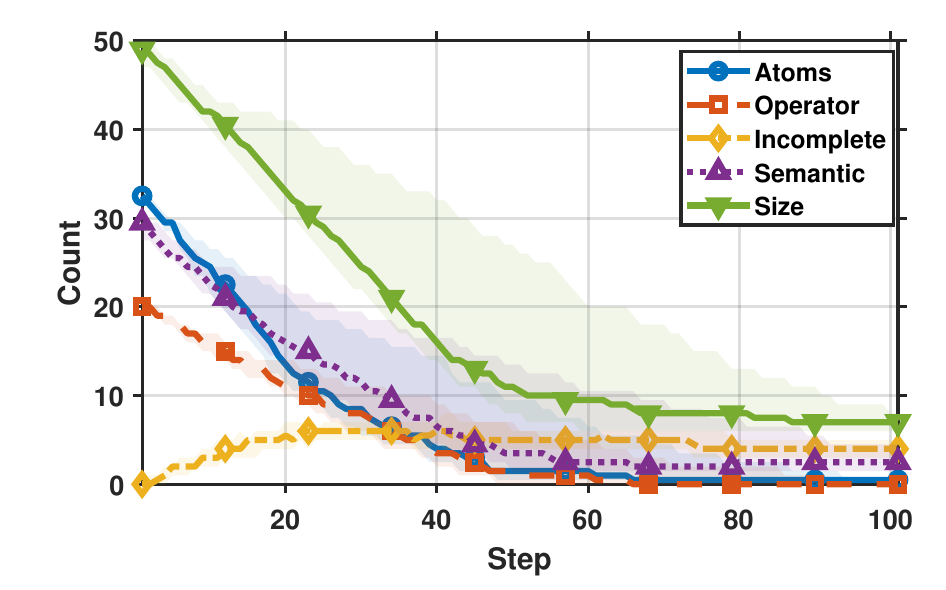}}
    \subfigure[Nguyen-11]{\includegraphics[width=.32\textwidth]{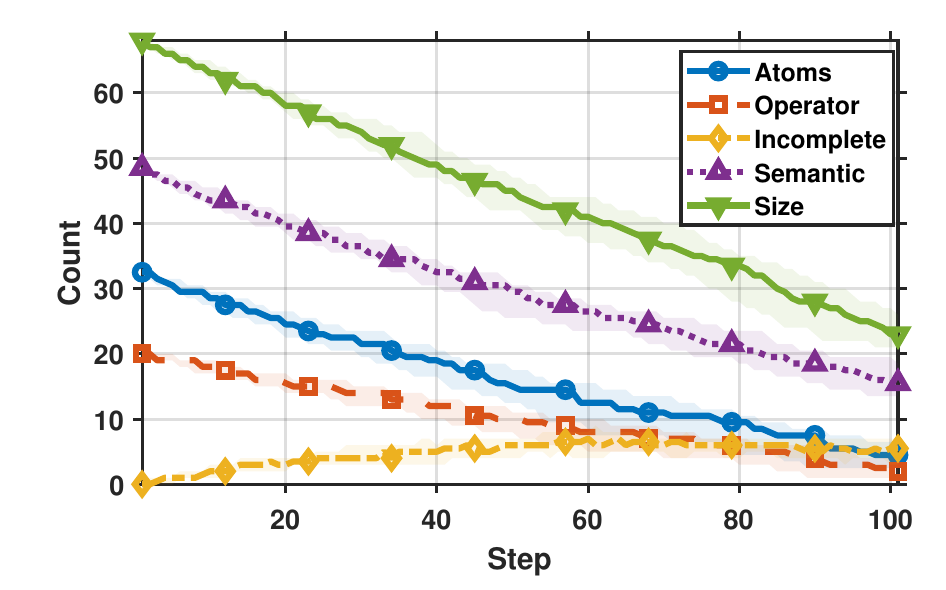}}

	\caption{Workspace composition during the final main derivation for Unsolvable benchmarks, showing the number of:
    total objects, and atomic, operator, incomplete and semantic objects.} 
	\label{fig:ws2}
\end{figure}

Figure \ref{fig:convergence2} shows the convergence plots for MGP on these problems,
while Figure \ref{fig:ws2} shows the workspace composition analysis, as was presented before
for the Solvable problems.
Results show that MGP struggles to construct viable approximations in these cases.
Standard training and testing performance over all runs are summarized in Table \ref{tab:test}.
Testing is done for extrapolation, considering the domain $[5,10]$
using the same number of points as those used for training.

\begin{figure}[t]
	\centering
	\subfigure[Nguyen-7]{\includegraphics[width=.99\textwidth]{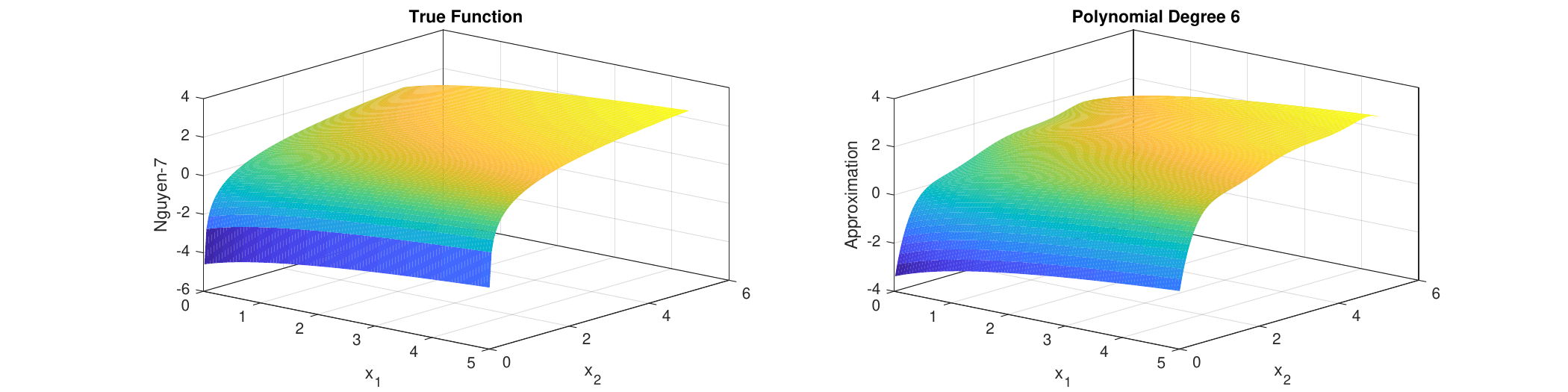}}
    \subfigure[Nguyen-11]{\includegraphics[width=.99\textwidth]{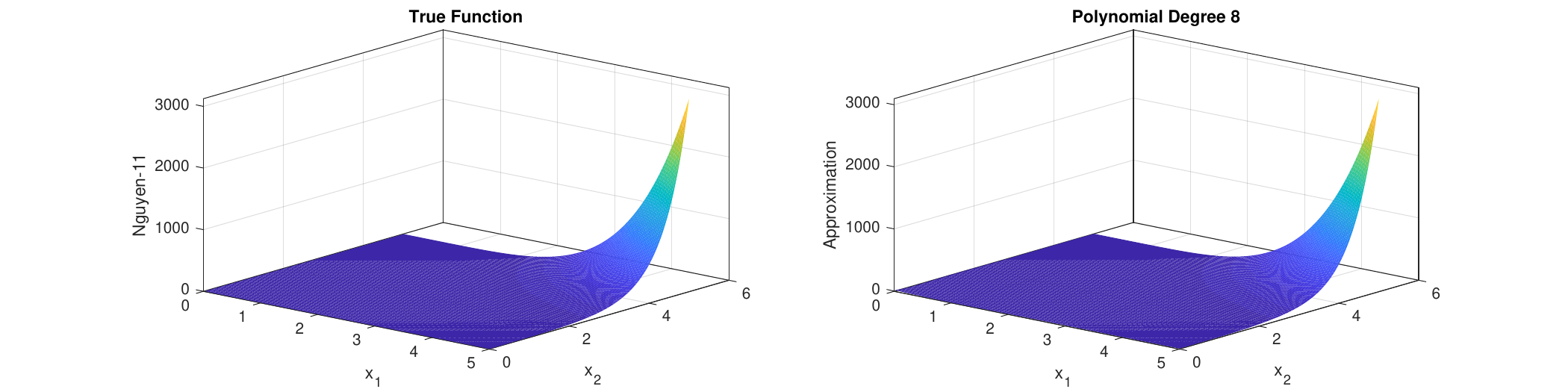}}
    \subfigure[Nguyen-8]{\includegraphics[width=.65\textwidth]{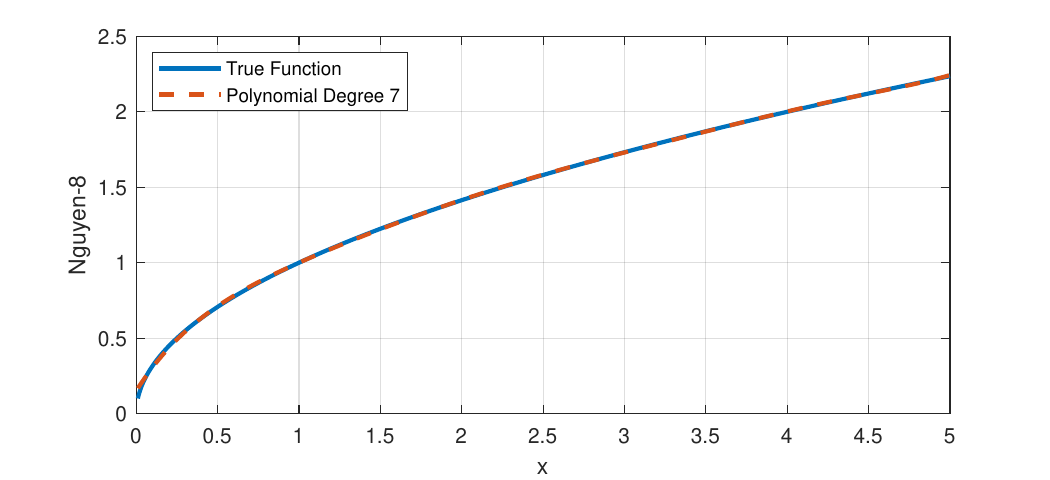}}

	\caption{Polynomial approximations for the Unsolvable benchmark problems.} 
	\label{fig:approx}
\end{figure}

These results are instructive for several reasons.
Consider that all three problems can be approximated quite well in the training domain using
a standard polynomial approximations, shown in the final columns of Table \ref{tab:test}.
The fit of the polynomial approximations are also depicted in Figure \ref{fig:approx}.
The models derived with MGP do not achieve the same approximation accuracy on the training domain, 
with median and even the best performance being an order of magnitude worse
compared to the polynomial on two problems (Nguyen-8 and Nguyen-11).
However, on the testing data, all models perform poorly, and the polynomial models
are actually much worse.
It is clear that the polynomial models are overfitted to the training data, as should be expected.
The MGP models are underfitted, and this seems to benefit them when the evaluation shifts to generalization.

However, as stated before, it would be trivial to add a numerical optimizer to MGP,
to fit implicit coefficients in the models as part of the semantic computation.
Since MGP has been shown to easily construct polynomial expressions (Nguyen-1 - Nguyen-4),
such a numerical optimizer would lead MGP to improve training performance, at the potential cost
of worse testing generalization (extrapolation), with the derivation process potentially focusing on polynomial approximations.
While this experimental analysis is left as as future work, these results illustrate
the importance of correctly choosing the lexicon for MGP, just like in any other SR system.
If not, approximate models are the best possible outcome, which are known to generalize poorly
outside the fitted domain.

\begin{table}[t]
    \centering
    \scriptsize
    \setlength{\tabcolsep}{4pt} 
    \caption{Performance of MGP on the Unsolvable benchmarks, showing the median and best solution over all 30 runs. Final columns show the degree and approximation error of the polynomial surrogate.}

    \begin{tabular}{lccccccccccc}
    \toprule
    \multirow{4}{*}{\textbf{Problem}}
        & \multicolumn{4}{c}{\textbf{MGP Train}}
        & \multicolumn{4}{c}{\textbf{MGP Test}}
        & \multicolumn{3}{c}{\textbf{Polynomial}} \\
    \cmidrule(lr){2-5}
    \cmidrule(lr){6-9}
    \cmidrule(lr){10-12}

        & \multicolumn{2}{c}{Median}
        & \multicolumn{2}{c}{Best}
        & \multicolumn{2}{c}{Median}
        & \multicolumn{2}{c}{Best}
        & \multirow{3}{*}{Degree}
        & \multicolumn{2}{c}{RMSE} \\

    \cmidrule(lr){2-3}
    \cmidrule(lr){4-5}
    \cmidrule(lr){6-7}
    \cmidrule(lr){8-9}
    \cmidrule(lr){11-12}

        & RMSE & $R^2$
        & RMSE & $R^2$
        & RMSE & $R^2$
        & RMSE & $R^2$
        & & Train & Test \\

    \midrule
    Nguyen-7
        & $1.14$ & $0.77$ & $0.29$ & $0.97$
        & $2.24$ & $-16.16$ & $1.58$ & $-7.4$
        & $6$ & $0.14$ & $7.6e+02$ \\

    Nguyen-8
        & $0.27$ & $0.70$ & $0.07$ & $0.97$
        & $0.74$ & $-0.07$ & $0.29$ & $0.84$
        & $7$ & $0.004$ & $2.0e+02$ \\

    Nguyen-11
        & $81.9$ & $0.89$ & $23.6$ & $0.96$
        & $2.1e+08$ & $-0.02$ & $5.3e+07$ & $0.00$
        & $8$ & $1.37$ & $6.9e+08$ \\

    \bottomrule
    \end{tabular}

    \label{tab:test}
\end{table}

\section{Discussion, Conclusions and Future Work}\label{sec:conclusions}

MGP presents an alternative formulation to the SR tasks, based on syntactic derivations
instead of the evolutionary search used in GP systems.
Like GP, MGP constructs hierarchical structures but appears to be more adapt
at discovering and reusing modular elements.
By constructing syntactic objects incrementally, MGP seems to avoid including unnecessary
elements, or at least construct models that are easy to simplify and
express in a compact way, unlike the bloated runs that are typical of classic GP.

Results indicate that the incremental tree-building process of MGP
is able to derive exact ground truth models, at least for the SR tasks evaluated here.
This is dependent, however, on properly choosing the atomic lexical items,
just like for other SR systems.
While these problems are relatively simple compared to more recent benchmarks \citep{srbench2},
they are nevertheless challenging for traditional GP systems \citep{Uy2010}.

Similar to GP, the biological inspiration based on minimalism might be implemented in other ways.
We do not claim that the particular implementation of MP principles in MGP are optimal in any way,
the goal was to show that these principles are relevant to program induction, and can
be brought to bare in problem domains like SR, what that looks like in the future requires additional work.
The importance of biological plausibility is not entirely evident,
a question often discussed in evolutionary computation.
Some of the most successful evolutionary methods are hybrid approaches,
where the ability to locate optimal solutions in complex search-spaces takes
precedent over implementing search mechanisms that align with evolutionary theory.
On the other hand, it may turn out that integrating more realistic evolutionary phenomena may
in fact be the key for the next generation of evolutionary algorithms,
including open-endedness, co-evolution, mating restrictions or developmental processes.
Similarly, our (overly) simplified implementation of the core principles of minimalist syntax,
while necessary to propose an initial baseline system, may be missing key elements
that are required to solve more challenging tasks.
These are open questions that require further study and analysis.
Building on the lessons learned from over 30 years of research since GP was first introduced, this work
outlines a novel research direction toward the development of automatic methods to derive symbolic computational models.

Future work will explore the following extensions, evaluations and improvements to MGP.
First, explore the role that internal $MERGE$ ($IM$) can potentially play in the proposed SR derivation process.
Second, integrate a numerical optimizer to allow MGP to properly fit model parameters, not just
model syntax and structure.
Third, extend the proposed system to other learning tasks, such as classification problems or
coding tasks.
Fourth, implement a parallel version of the iterative derivation process for improved efficiency.
Finally, evaluate MGP on a broader set of tasks, including other synthetic problems with known
ground truth models and black-box regression tasks.
One particular application domain of interest is medical applications, where model
clarity and interpretability is of paramount importance \citep{Sakallioglu2026}.


\section*{Acknowledgments}
The first author was supported by TecNM project 24550.26-P and SECIHTI (Mexico) project CF-2023-I-724.

\bibliographystyle{apalike}
\bibliography{ecjsample}

@article{koza10,
  year = {2010},
  month = may,
  publisher = {Springer Science and Business Media {LLC}},
  volume = {11},
  number = {3-4},
  pages = {251--284},
  author = {John R. Koza},
  title = {Human-competitive results produced by genetic programming},
  journal = {Genetic Programming and Evolvable Machines}
}

@BOOK{koza,
  title     = "Genetic programming",
  author    = "Koza, John R",
  publisher = "Bradford Books",
  series    = "Complex Adaptive Systems",
  month     =  dec,
  year      =  1992,
  address   = "Cambridge, MA"
}

@book{mitchell_artificial_2019,
	address = {New York},
	edition = {First edition.},
	title = {Artificial intelligence: a guide for thinking humans},
	isbn = {978-0-374-25783-5},
	shorttitle = {Artificial intelligence},
	language = {eng},
	publisher = {Farrar, Straus and Giroux},
	author = {Mitchell, Melanie},
	year = {2019},
}

@techreport{mccarthy1955proposal,
  title={A proposal for the {D}artmouth summer research project on artificial intelligence},
  author={McCarthy, John and Minsky, Marvin L and Rochester, Nathaniel and Shannon, Claude E},
  year={1955},
  month={Aug},
  institution={Dartmouth College},
  address={Hanover, NH},
  note={Available at \url{http://jmc.stanford.edu/articles/dartmouth.html}}
}

@article{Stork2020,
  title = {A new taxonomy of global optimization algorithms},
  volume = {21},
  ISSN = {1572-9796},
  url = {http://dx.doi.org/10.1007/s11047-020-09820-4},
  DOI = {10.1007/s11047-020-09820-4},
  number = {2},
  journal = {Natural Computing},
  publisher = {Springer Science and Business Media LLC},
  author = {Stork,  J\"{o}rg and Eiben,  A. E. and Bartz-Beielstein,  Thomas},
  year = {2020},
  month = Nov,
  pages = {219–242}
}

@inbook{Srensen2018,
  title = {A History of Metaheuristics},
  ISBN = {9783319071534},
  url = {http://dx.doi.org/10.1007/978-3-319-07153-4_4-1},
  DOI = {10.1007/978-3-319-07153-4_4-1},
  booktitle = {Handbook of Heuristics},
  publisher = {Springer International Publishing},
  author = {S\"{o}rensen,  Kenneth and Sevaux,  Marc and Glover,  Fred},
  year = {2018},
  pages = {1–18}
}

@article{Mengistu2016,
  title = {The Evolutionary Origins of Hierarchy},
  volume = {12},
  ISSN = {1553-7358},
  url = {http://dx.doi.org/10.1371/journal.pcbi.1004829},
  DOI = {10.1371/journal.pcbi.1004829},
  number = {6},
  journal = {PLOS Computational Biology},
  publisher = {Public Library of Science (PLoS)},
  author = {Mengistu,  Henok and Huizinga,  Joost and Mouret,  Jean-Baptiste and Clune,  Jeff},
  editor = {Sporns,  Olaf},
  year = {2016},
  month = June,
  pages = {e1004829}
}

@book{Kronberger2024,
  title = {Symbolic Regression},
  ISBN = {9781315166407},
  url = {http://dx.doi.org/10.1201/9781315166407},
  DOI = {10.1201/9781315166407},
  publisher = {Chapman and Hall/CRC},
  author = {Kronberger,  Gabriel and Burlacu,  Bogdan and Kommenda,  Michael and Winkler,  Stephan M. and Affenzeller,  Michael},
  year = {2024},
  month = July 
}

@article{Atzmueller2024,
  title = {Explainable and interpretable machine learning and data mining},
  volume = {38},
  ISSN = {1573-756X},
  url = {http://dx.doi.org/10.1007/s10618-024-01041-y},
  DOI = {10.1007/s10618-024-01041-y},
  number = {5},
  journal = {Data Mining and Knowledge Discovery},
  publisher = {Springer Science and Business Media LLC},
  author = {Atzmueller,  Martin and F\"{u}rnkranz,  Johannes and Kliegr,  Tomáš and Schmid,  Ute},
  year = {2024},
  month = jul,
  pages = {2571–2595}
}

@article{romera2024mathematical,
  title={Mathematical discoveries from program search with large language models},
  author={Romera-Paredes, Bernardino and Barekatain, Mohammadamin and Novikov, Alexander and Balog, Matej and Kumar, M Pawan and Dupont, Emilien and Ruiz, Francisco JR and Ellenberg, Jordan S and Wang, Pengming and Fawzi, Omar and others},
  journal={Nature},
  volume={625},
  number={7995},
  pages={468--475},
  year={2024},
  publisher={Nature Publishing Group UK London}
}

@book{Uriagereka2000,
  title = {Rhyme and Reason: An Introduction to Minimalist Syntax},
  ISBN = {9780262285377},
  url = {http://dx.doi.org/10.7551/mitpress/5949.001.0001},
  DOI = {10.7551/mitpress/5949.001.0001},
  publisher = {The MIT Press},
  author = {Uriagereka,  Juan},
  year = {2000}
}

@article{Castelli2023,
  title = {Commentary for the GPEM peer commentary special section on W. B. Langdon’s “Jaws 30”},
  volume = {24},
  ISSN = {1573-7632},
  url = {http://dx.doi.org/10.1007/s10710-023-09468-w},
  DOI = {10.1007/s10710-023-09468-w},
  number = {2},
  journal = {Genetic Programming and Evolvable Machines},
  publisher = {Springer Science and Business Media LLC},
  author = {Castelli,  Mauro},
  year = {2023},
  month = Nov 
}

@inproceedings{gleam,
author = {Saini, Anil Kumar and Spector, Lee},
title = {GLEAM: genetic learning by extraction and absorption of modules},
year = {2021},
isbn = {9781450383516},
publisher = {Association for Computing Machinery},
address = {New York, NY, USA},
url = {https://doi.org/10.1145/3449726.3459544},
doi = {10.1145/3449726.3459544},
booktitle = {Proceedings of the Genetic and Evolutionary Computation Conference Companion},
pages = {263–264},
numpages = {2},
location = {Lille, France},
series = {GECCO '21}
}

@book{dawkins1976selfish,
  title={The Selfish Gene},
  author={Dawkins, Richard},
  year={1976},
  publisher={Oxford University Press}
}

@article{Silva2011,
  title = {Operator equalisation for bloat free genetic programming and a survey of bloat control methods},
  volume = {13},
  ISSN = {1573-7632},
  url = {http://dx.doi.org/10.1007/s10710-011-9150-5},
  DOI = {10.1007/s10710-011-9150-5},
  number = {2},
  journal = {Genetic Programming and Evolvable Machines},
  publisher = {Springer Science and Business Media LLC},
  author = {Silva,  Sara and Dignum,  Stephen and Vanneschi,  Leonardo},
  year = {2011},
  month = Nov,
  pages = {197–238}
}

@inbook{Langdon1998,
  title = {Fitness Causes Bloat},
  ISBN = {9781447104278},
  url = {http://dx.doi.org/10.1007/978-1-4471-0427-8_2},
  DOI = {10.1007/978-1-4471-0427-8_2},
  booktitle = {Soft Computing in Engineering Design and Manufacturing},
  publisher = {Springer London},
  author = {Langdon,  W. B. and Poli,  R.},
  year = {1998},
  pages = {13–22}
}

@book{Chomsky1995,
  title     = {The Minimalist Program},
  author    = {Chomsky, Noam},
  year      = {1995},
  publisher = {The MIT Press},
  address   = {Cambridge, MA},
  series    = {Current Studies in Linguistics},
  volume    = {28}
}

@book{berwick2016why,
  title={Why Only Us: Language and Evolution},
  author={Berwick, Robert C and Chomsky, Noam},
  year={2016},
  publisher={MIT Press},
  address={Cambridge, MA}
}

@article{Fitch2014,
  title = {Toward a computational framework for cognitive biology: Unifying approaches from cognitive neuroscience and comparative cognition},
  volume = {11},
  ISSN = {1571-0645},
  url = {http://dx.doi.org/10.1016/j.plrev.2014.04.005},
  DOI = {10.1016/j.plrev.2014.04.005},
  number = {3},
  journal = {Physics of Life Reviews},
  publisher = {Elsevier BV},
  author = {Fitch,  W. Tecumseh},
  year = {2014},
  month = Sept,
  pages = {329–364}
}

@incollection{Fitch2017,
  author    = {Fitch, W. Tecumseh},
  title     = {Dendrophilia and the Evolution of Syntax},
  booktitle = {Origins of Human Language: Continuities and Discontinuities with Nonhuman Primates},
  editor    = {Bo{\"{e}}, Louis-Jean and Fagot, Jo{\"{e}}l and Perrier, Pascal and Schwartz, Jean-Luc},
  publisher = {Peter Lang},
  address   = {Bern},
  year      = {2017},
  pages     = {305--328},
  doi       = {10.3726/b12405},
  isbn      = {9783631737262}
}

@article{Uy2010,
  title = {Semantically-based crossover in genetic programming: application to real-valued symbolic regression},
  volume = {12},
  ISSN = {1573-7632},
  url = {http://dx.doi.org/10.1007/s10710-010-9121-2},
  DOI = {10.1007/s10710-010-9121-2},
  number = {2},
  journal = {Genetic Programming and Evolvable Machines},
  publisher = {Springer Science and Business Media LLC},
  author = {Uy,  Nguyen Quang and Hoai,  Nguyen Xuan and O’Neill,  Michael and McKay,  R. I. and Galván-López,  Edgar},
  year = {2010},
  month = July,
  pages = {91–119}
}

@inbook{Trujillo2018,
  title = {Local Search is Underused in Genetic Programming},
  ISBN = {9783319970882},
  ISSN = {1932-0167},
  url = {http://dx.doi.org/10.1007/978-3-319-97088-2_8},
  DOI = {10.1007/978-3-319-97088-2_8},
  booktitle = {Genetic Programming Theory and Practice XIV},
  publisher = {Springer International Publishing},
  author = {Trujillo,  Leonardo and Z-Flores,  Emigdio and Juárez-Smith,  Perla S. and Legrand,  Pierrick and Silva,  Sara and Castelli,  Mauro and Vanneschi,  Leonardo and Sch\"{u}tze,  Oliver and Muñoz,  Luis},
  year = {2018},
  pages = {119–137}
}

@book{lucretius2007nature,
  title={The Nature of Things},
  author={Lucretius, Titus Carus},
  translator={Stallings, A. E.},
  year={2007},
  publisher={Penguin Books},
  address={London},
  note={Translated with an introduction by A.E. Stallings}
}

@article{ShwartzZiv2022,
  title = {Tabular data: Deep learning is not all you need},
  volume = {81},
  ISSN = {1566-2535},
  url = {http://dx.doi.org/10.1016/j.inffus.2021.11.011},
  DOI = {10.1016/j.inffus.2021.11.011},
  journal = {Information Fusion},
  publisher = {Elsevier BV},
  author = {Shwartz-Ziv,  Ravid and Armon,  Amitai},
  year = {2022},
  month = May,
  pages = {84–90}
}

@book{chomsky1965aspects,
  title={Aspects of the Theory of Syntax},
  author={Chomsky, Noam},
  year={1965},
  publisher={MIT Press},
  address={Cambridge, MA},
  series={Special Technical Report of the Research Laboratory of Electronics of the Massachusetts Institute of Technology},
  number={11}
}

@article{Pan2024,
  title = {Introduction: workspace,  MERGE and labelling},
  volume = {41},
  ISSN = {1613-3676},
  url = {http://dx.doi.org/10.1515/tlr-2024-2001},
  DOI = {10.1515/tlr-2024-2001},
  number = {1},
  journal = {The Linguistic Review},
  publisher = {Walter de Gruyter GmbH},
  author = {Pan,  Victor Junnan and Saito,  Mamoru and Du,  Yuqiao},
  year = {2024},
  month = Jan,
  pages = {1–5}
}

@article{Matsumoto2023,
  title = {Syntactic theory of mathematical expressions},
  volume = {146},
  ISSN = {0010-0285},
  url = {http://dx.doi.org/10.1016/j.cogpsych.2023.101606},
  DOI = {10.1016/j.cogpsych.2023.101606},
  journal = {Cognitive Psychology},
  publisher = {Elsevier BV},
  author = {Matsumoto,  Daiki and Nakai,  Tomoya},
  year = {2023},
  month = Nov,
  pages = {101606}
}

@article{Trotzke2020,
  title = {Constructions in Minimalism: A Functional Perspective on Cyclicity},
  volume = {11},
  ISSN = {1664-1078},
  url = {http://dx.doi.org/10.3389/fpsyg.2020.02152},
  DOI = {10.3389/fpsyg.2020.02152},
  journal = {Frontiers in Psychology},
  publisher = {Frontiers Media SA},
  author = {Trotzke,  Andreas},
  year = {2020},
  month = Sept 
}

@incollection{chomsky2004beyond,
title={Beyond Explanatory Adequacy},
author={Chomsky, Noam},
booktitle={Structures and Beyond: The Cartography of Syntactic Structures, Volume 3},
editor={Belletti, Adriana},
pages={104--131},
year={2004},
publisher={Oxford University Press},
address={Oxford}}

@article{komachi2019generative,
  title={Generative procedure revisited},
  author={Komachi, Masayuki and Kitahara, Hisatsugu and Uchibori, Asako and Takita, Kensuke},
  journal={Reports of the Keio Institute of Cultural and Linguistic Studies},
  volume={50},
  pages={269--283},
  year={2019},
  publisher={慶應義塾大学言語文化研究所}
}

@book{chomsky1957syntactic,
  author    = {Chomsky, Noam},
  title     = {Syntactic Structures},
  year      = {1957},
  publisher = {Mouton},
  address   = {The Hague}
}

@incollection{marantz1997no,
  author    = {Alec Marantz},
  title     = {No Escape from Syntax: Don't Try Morphological Analysis in the Privacy of Your Own Lexicon},
  booktitle = {Proceedings of the 21st Annual Penn Linguistics Colloquium},
  editor    = {Alexis Dimitriadis and Laura Siegel and Clarissa Surek-Clark and Alexander Williams},
  series    = {Penn Working Papers in Linguistics},
  volume    = {4},
  number    = {2},
  pages     = {201--225},
  year      = {1997},
  publisher = {University of Pennsylvania}
}

@article{starke2010nanosyntax,
  title={Nanosyntax: A short primer to a new approach to language},
  author={Starke, Michal},
  journal={Nordlyd},
  volume={36},
  number={1},
  pages={1--6},
  year={2010},
  url={https://septentrio.uit.no/index.php/nordlyd/article/view/213}
}

@book{chomsky2023merge,
  title = {Merge and the Strong Minimalist Thesis},
  ISBN = {9781009343268},
  url = {http://dx.doi.org/10.1017/9781009343244},
  DOI = {10.1017/9781009343244},
  publisher = {Cambridge University Press},
  author = {Chomsky,  Noam and Seely,  T. Daniel and Berwick,  Robert C. and Fong,  Sandiway and Huybregts,  M. A. C. and Kitahara,  Hisatsugu and McInnerney,  Andrew and Sugimoto,  Yushi},
  year = {2023},
  month = Nov 
}

@book{ONeill2003,
  title = {Grammatical Evolution},
  ISBN = {9781461504474},
  url = {http://dx.doi.org/10.1007/978-1-4615-0447-4},
  DOI = {10.1007/978-1-4615-0447-4},
  publisher = {Springer US},
  author = {O’Neill,  Michael and Ryan,  Conor},
  year = {2003}
}

@article{McKay2010,
  title = {Grammar-based Genetic Programming: a survey},
  volume = {11},
  ISSN = {1573-7632},
  url = {http://dx.doi.org/10.1007/s10710-010-9109-y},
  DOI = {10.1007/s10710-010-9109-y},
  number = {3-4},
  journal = {Genetic Programming and Evolvable Machines},
  publisher = {Springer Science and Business Media LLC},
  author = {McKay,  Robert I. and Hoai,  Nguyen Xuan and Whigham,  Peter Alexander and Shan,  Yin and O’Neill,  Michael},
  year = {2010},
  month = May,
  pages = {365–396}
}

@book{darwin1871descent,
  title={The Descent of Man, and Selection in Relation to Sex},
  author={Darwin, Charles},
  year={1871},
  publisher={John Murray},
  address={London},
  volume={1--2}
}

@article{Aranha2021,
  title = {Metaphor-based metaheuristics, a call for action: the elephant in the room},
  volume = {16},
  ISSN = {1935-3820},
  url = {http://dx.doi.org/10.1007/s11721-021-00202-9},
  DOI = {10.1007/s11721-021-00202-9},
  number = {1},
  journal = {Swarm Intelligence},
  publisher = {Springer Science and Business Media LLC},
  author = {Aranha,  Claus and Camacho Villalón,  Christian L. and Campelo,  Felipe and Dorigo,  Marco and Ruiz,  Rubén and Sevaux,  Marc and S\"{o}rensen,  Kenneth and St\"{u}tzle,  Thomas},
  year = {2021},
  month = Nov,
  pages = {1–6}
}

@article{Srensen2013,
  title = {Metaheuristics—the metaphor exposed},
  volume = {22},
  ISSN = {1475-3995},
  url = {http://dx.doi.org/10.1111/itor.12001},
  DOI = {10.1111/itor.12001},
  number = {1},
  journal = {International Transactions in Operational Research},
  publisher = {Wiley},
  author = {S\"{o}rensen,  Kenneth},
  year = {2013},
  month = Feb,
  pages = {3–18}
}

@book{stanley2015greatness,
  title={Why Greatness Cannot Be Planned: The Myth of the Objective},
  author={Stanley, Kenneth O. and Lehman, Joel},
  year={2015},
  publisher={Springer},
  address={Heidelberg, Germany},
  isbn={978-3-319-15523-4}
}

@article{stanley2002evolving,
  title={Evolving neural networks through augmenting topologies},
  author={Stanley, Kenneth O and Miikkulainen, Risto},
  journal={Evolutionary Computation},
  volume={10},
  number={2},
  pages={99--127},
  year={2002},
  publisher={MIT Press}
}

@inproceedings{Miller2000CartesianGP,
author = {Miller, Julian F. and Thomson, Peter},
title = {Cartesian Genetic Programming},
year = {2000},
isbn = {3540673393},
publisher = {Springer-Verlag},
address = {Berlin, Heidelberg},
booktitle = {Proceedings of the European Conference on Genetic Programming},
pages = {121–132},
numpages = {12}
}

@book{linearGP,
  author    = {Markus Brameier and Wolfgang Banzhaf},
  title     = {Linear Genetic Programming},
  publisher = {Springer},
  address   = {New York},
  year      = {2007},
  isbn      = {978-0-387-31030-5},
  doi       = {10.1007/978-0-387-31030-5}
}

@article{KUPPURAJU1985,
  title = {HIERARCHICAL DECISION MAKING IN SYSTEM DESIGN},
  volume = {8},
  ISSN = {1029-0273},
  url = {http://dx.doi.org/10.1080/03052158508902491},
  DOI = {10.1080/03052158508902491},
  number = {3},
  journal = {Engineering Optimization},
  publisher = {Informa UK Limited},
  author = {Kuppuraju,  N. and GANESAN,  S. and MISTREE,  F. and SOBIESKI,  J. S.},
  year = {1985},
  month = Jan,
  pages = {223–252}
}

@inbook{Sakallioglu2026,
  title = {Combining Grammatical Evolution with LLM-based Local Search to Improve Interpretability},
  ISBN = {9783032236074},
  ISSN = {1611-3349},
  url = {http://dx.doi.org/10.1007/978-3-032-23607-4_14},
  DOI = {10.1007/978-3-032-23607-4_14},
  booktitle = {Applications of Evolutionary Computation},
  publisher = {Springer Nature Switzerland},
  author = {Sakallioglu,  Berfin and Santos,  Frederico J. J. B. and Parra,  Daniel and Qiu,  Yuxin and Nicolau,  Miguel and Trujillo,  Leonardo and Hidalgo,  J. Ignacio},
  year = {2026},
  pages = {217–233}
}

@inproceedings{srbench2,
 author = {La Cava, William and Orzechowski, Patryk and Burlacu, Bogdan and de Franca, Fabricio and Virgolin, Marco and Jin, Ying and Kommenda, Michael and Moore, Jason},
 booktitle = {Proceedings of the Neural Information Processing Systems Track on Datasets and Benchmarks},
 editor = {J. Vanschoren and S. Yeung},
 pages = {},
 publisher = {Curran},
 title = {Contemporary Symbolic Regression Methods and their Relative Performance},
 volume = {1},
 year = {2021},
}

@article{rudin1,
  year = {2019},
  month = may,
  publisher = {Springer Science and Business Media {LLC}},
  volume = {1},
  number = {5},
  pages = {206--215},
  author = {Cynthia Rudin},
  title = {Stop explaining black box machine learning models for high stakes decisions and use interpretable models instead},
  journal = {Nature Machine Intelligence}
}

@inproceedings{virgolin,
author = {Virgolin, Marco and De Lorenzo, Andrea and Medvet, Eric and Randone, Francesca},
title = {Learning a Formula of Interpretability to Learn Interpretable Formulas},
year = {2020},
isbn = {978-3-030-58114-5},
publisher = {Springer-Verlag},
address = {Berlin, Heidelberg},
url = {https://doi.org/10.1007/978-3-030-58115-2_6},
doi = {10.1007/978-3-030-58115-2_6},
booktitle = {Parallel Problem Solving from Nature – PPSN XVI: 16th International Conference, PPSN 2020, Leiden, The Netherlands, September 5-9, 2020, Proceedings, Part II},
pages = {79–93},
numpages = {15},
location = {Leiden, The Netherlands}
}

@article{push,
  title={Genetic programming and autoconstructive evolution with the Push programming language},
  author={Spector, Lee and Robinson, Alan},
  journal={Genetic Programming and Evolvable Machines},
  volume={3},
  number={1},
  pages={7--40},
  year={2002},
  publisher={Springer},
  doi={10.1023/A:1014538503543}
}

@inproceedings{gsgp,
 author = {Moraglio, Alberto and Krawiec, Krzysztof and Johnson, Colin G.},
 title = {Geometric Semantic Genetic Programming},
 booktitle = {Proceedings of the 12th International Conference on Parallel Problem Solving from Nature - Volume Part I},
 series = {PPSN'12},
 year = {2012},
 location = {Taormina, Italy},
 pages = {21--31},
 numpages = {11},
}

@incollection{ffx,
  year = {2011},
  publisher = {Springer New York},
  pages = {235--260},
  author = {Trent McConaghy},
  title = {{FFX}: Fast,  Scalable,  Deterministic Symbolic Regression Technology},
  booktitle = {Genetic and Evolutionary Computation}
}

@misc{feat,
      title={Learning concise representations for regression by evolving networks of trees}, 
      author={William La Cava and Tilak Raj Singh and James Taggart and Srinivas Suri and Jason H. Moore},
      year={2019},
      eprint={1807.00981},
      archivePrefix={arXiv},
      primaryClass={cs.NE}
}

@inproceedings{kaizen,
  series = {GECCO ’14},
  title = {Kaizen programming},
  url = {http://dx.doi.org/10.1145/2576768.2598264},
  DOI = {10.1145/2576768.2598264},
  booktitle = {Proceedings of the 2014 Annual Conference on Genetic and Evolutionary Computation},
  publisher = {ACM},
  author = {De Melo,  Vinícius Veloso},
  year = {2014},
  month = July,
  pages = {895–902},
  collection = {GECCO ’14}
}

@article{random,
  year = {2018},
  month = feb,
  publisher = {Springer Science and Business Media {LLC}},
  volume = {11},
  number = {1},
  author = {Moshe Sipper and Weixuan Fu and Karuna Ahuja and Jason H. Moore},
  title = {Investigating the parameter space of evolutionary algorithms},
  journal = {{BioData} Mining}
}

@article{perla,
  year = {2019},
  month = mar,
  publisher = {Springer Science and Business Media {LLC}},
  volume = {20},
  number = {3},
  pages = {351--384},
  author = {Perla Ju{\'{a}}rez-Smith and Leonardo Trujillo and Mario Garc{\'{\i}}a-Valdez and Francisco Fern{\'{a}}ndez de Vega and Francisco Ch{\'{a}}vez},
  title = {Local search in speciation-based bloat control for genetic programming},
  journal = {Genetic Programming and Evolvable Machines}
}

@Article{m5gp,
AUTHOR = {Cárdenas Florido, Luis and Trujillo, Leonardo and Hernandez, Daniel E. and Muñoz Contreras, Jose Manuel},
TITLE = {{M5GP}: Parallel Multidimensional Genetic Programming with Multidimensional Populations for Symbolic Regression},
JOURNAL = {Mathematical and Computational Applications},
VOLUME = {29},
YEAR = {2024},
NUMBER = {2},
ARTICLE-NUMBER = {25},
URL = {https://www.mdpi.com/2297-8747/29/2/25},
ISSN = {2297-8747},
}

@article{new2,
  title = {Genetic Programming for Automatically Evolving Multiple Features to Classification},
  ISSN = {1530-9304},
  DOI = {10.1162/evco_a_00359},
  journal = {Evolutionary Computation},
  publisher = {MIT Press},
  author = {Wang,  Peng and Xue,  Bing and Liang,  Jing and Zhang,  Mengjie},
  year = {2024},
  month = nov,
  pages = {1–28}
}

\end{document}